\documentclass[lettersize,journal]{IEEEtran}
\usepackage{array}
\usepackage[caption=false,font=normalsize,labelfont=sf,textfont=sf]{subfig}
\usepackage{textcomp}
\usepackage{stfloats}
\usepackage{url}
\usepackage{verbatim}
\usepackage{graphicx}
\usepackage{cite}
\usepackage{amsthm}
\usepackage{amsmath,amsfonts}
\usepackage{amssymb}
\usepackage{threeparttable}
\usepackage{booktabs}
\usepackage{multicol}
\usepackage{multirow, bm}
\usepackage{makecell}
\usepackage{graphicx}
\usepackage{float}

\usepackage{caption}
\usepackage{subfloat}

\theoremstyle{plain}
\newtheorem{theorem}{Theorem}[section]

\theoremstyle{definition}
\newtheorem{definition}[theorem]{Definition}

\theoremstyle{remark}

\usepackage{algorithm}
\usepackage{algpseudocode}

\usepackage[colorlinks,
            linkcolor=blue,       
            anchorcolor=blue,  
            citecolor=blue,        
            ]{hyperref}

\renewcommand{\footnoterule}{%
\kern -3pt 
\hrule width 2in height 0.4pt 
\kern 2.6pt} 
            
\begin{document}

\title{DEER: A Delay-Resilient Framework for Reinforcement Learning  with Variable Delays }

\author{
\IEEEauthorblockN{Bo Xia\IEEEauthorrefmark{1}, Yilun Kong\IEEEauthorrefmark{1}\thanks{\IEEEauthorrefmark{1}Equal Contribution}, Yongzhe Chang, Bo Yuan, Zhiheng Li\IEEEauthorrefmark{2}, Xueqian Wang\IEEEauthorrefmark{2}\thanks{\IEEEauthorrefmark{2}Corresponding Author}, Bin Liang}\\
\IEEEauthorblockA{\textit{Tsinghua University, Shenzhen International Graduate School} \\
Shenzhen, China}
}

\maketitle

\begin{abstract}
Classic reinforcement learning (RL) frequently confronts challenges in tasks involving delays, which cause a mismatch between received observations and subsequent actions, thereby deviating from the Markov assumption.
Existing methods usually tackle this issue with end-to-end solutions using state augmentation. 
However, these black-box approaches often involve incomprehensible processes and redundant information in the information states, causing instability and potentially undermining the overall performance.
To alleviate the delay challenges in RL, we propose \textbf{DEER (\underline{D}elay-resilient \underline{E}ncoder-\underline{E}nhanced \underline{R}L)}, a framework designed to effectively enhance the interpretability and address the random delay issues.
DEER employs a pretrained encoder to map delayed states, along with their variable-length past action sequences resulting from different delays, into hidden states, which is trained on delay-free environment datasets.
In a variety of delayed scenarios, the trained encoder can seamlessly integrate with standard RL algorithms without requiring additional modifications and enhance the delay-solving capability by simply adapting the input dimension of the original algorithms.
 We evaluate DEER through extensive experiments on Gym and Mujoco environments. 
 The results confirm that DEER is superior to state-of-the-art RL algorithms in both constant and random delay settings.
\end{abstract}


\section{Introduction}

\IEEEPARstart{D}{eep} reinforcement learning has made substantial advancements in games \cite{1-games-1, 1-games-2} and large language models \cite{1-LLM-1, 1-LLM-2}, where most research assumes that action execution and state observation occur instantaneously. 
However, delays are inevitable in real-world tasks such as robotics \cite{1-robot-1, 1-robot-2}, remote control \cite{1-remote-1} and distributed communication \cite{1-commu-1}.
Prior research \cite{1-previous-1, 1-previous-2} has revealed the substantial impact of delays on an agent's decision process, leading to not only performance degradation but also potential instability in dynamic systems, which poses severe risks in real-world applications. 
In particular, in self-driving scenarios, even minor delays in the observation and execution modules can markedly increase the risk of accidents.

\begin{figure*}[ht]
    \centering
    \includegraphics[width=0.7\linewidth]{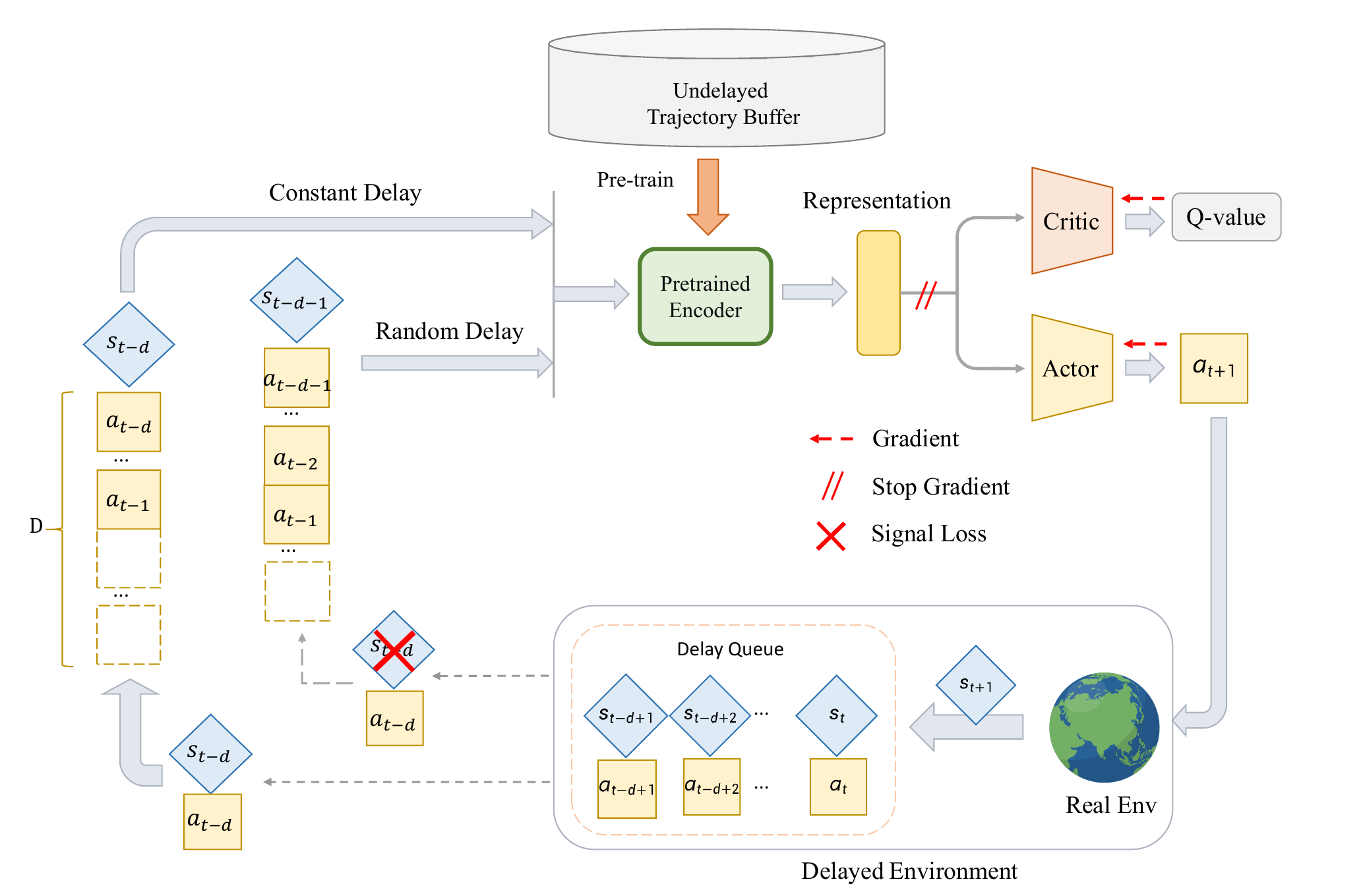}
    \caption{
    \textbf{Overview of DEER.}  The overall process consists of two main parts: pre-training an encoder using the offline dataset from undelayed environments to obtain a fixed-length feature representation for information states, and utilizing these context representations to guide decision-making during the agent's interaction with the delayed environments. Specifically, for policy learning in environments with a constant delay of $d$, the construction of the information is depicted by the thick solid line. In random environments, where the agent misses state $s_{t-d}$,  
    the thin dashed line illustrates the process of constructing the information state using the preceding state and action sequences from the previous time step.
    The variable $\text{D}$ in the figure denotes the maximum delay the agent can tolerate ($\text{D} = d_I + d_M$). Subsequently, information states are mapped into fixed-length context representations, which the agent uses for decision-making.
    }
    \label{fig1}
\end{figure*}

Despite the ubiquity of delays as a practical challenge, related research in the domain of RL remains scarce. Existing methods largely fall into two categories: model-free and model-based approaches.
Most model-free approaches~\cite{2-modelfree-1,2-modelfree-2,2-modelfree-3,2-modelfree-4,2-modelfree-5,2-modelfree-6,2-modelfree-7} rely on information states (consisting of delayed states and their corresponding action sequences) to transform delayed MDPs into equivalent undelayed ones.
Although these methods have achieved some success, their effectiveness is limited by the information state space's dimension. 
On the one hand, fixed input dimension methods are tailored for environments with constant delays, making them unsuitable for new tasks with different or random delays. 
On the other hand, the dimension of the information state space grows linearly with the length of delay, leading to exponential computational requirements and suboptimal policy learning.
By contrast, model-based methods \cite{2-modelbased-1,2-modelbased-2,2-modelbased-3,2-modelbased-4,2-modelbased-5} aim to predict the current state using the agent's recently received delayed state and action sequence. 
While being effective in static contexts, their adaptability in nondeterministic environments requires further enhancement. 
Some typical examples include a predictive model using unrolled Gated Recurrent Unit (GRU) \cite{GRU} modules to iteratively generate a single action \cite{2-modelbased-4}, and the Delayed-Q algorithm for making decisions based on iterative forward dynamic predictions \cite{2-modelbased-5}. 
However, both methods suffer from  issues including inference time, model precision, and cumulative errors, which can significantly impact their overall performance.

Considering the outlined challenges, we propose \textbf{\underline{D}elay-resilient \underline{E}ncoder-\underline{E}nhanced \underline{R}L (DEER)}, which leverages an encoder pretrained on offline datasets to enhance online learning in delayed environments.
Instead of making direct decisions using information states, we initially map these states into a hidden space known as the context representation space. 
Actions are subsequently inferred based on these context representations. 
The overview of DEER is shown in Fig.\ref{fig1}. 

It is worth noting that both the encoder and decoder modules are trained simultaneously in the pre-training phase, and only the encoder module is used during the decision-making phase to generate a context representation with a semantic embedding of the information state.
This embedding encapsulates the implicit information about both the current state and historical states, effectively serving as a high-dimensional state representation without delay, which can be directly used by standard RL algorithms to generate the current action. 
This process features three key advantages: 
(1) The trained encoder can easily generalize to diverse delay environments, whether constant or random, as it transforms the information state into a fixed-length vector. This eliminates the need to modify the agent's structure for different delay scenarios; 
(2) The proposed approach explicitly breaks down the end-to-end decision process into two distinct stages: encoding the information state and making decisions based on the embedding, which significantly improves the interpretability of the entire process (further discussion is provided in Appendix \ref{interpretability});
(3) The method enhances robustness compared to model-based approaches, as the latter use predicted states while DEER uses latent representation (additional analysis is presented in Appendix \ref{robustness}).

Specifically, considering the scarcity and costliness of expert data in practical settings, we pretrain the encoder-decoder model using an undelayed offline dataset primarily composed of random trajectories, supplemented with a limited number of expert trajectories.
Furthermore, DEER can be seamlessly integrated with any standard RL algorithm. 
In this paper, we employ Soft Actor-Critic (SAC) \cite{haarnoja2018soft} as the decision module, and comprehensive experiments on Gym and Mujoco confirm that our approach is superior to state-of-the-art methods in both constant and random delay environments.

The main contributions of this paper are summarized as follows: 

\begin{itemize}
    \item \textbf{Innovative use of offline datasets:} DEER leverages offline datasets from delay-free environment tasks to handle tasks occurring in delayed environments.
    \item \textbf{Versatile framework:} DEER enhances agent performance in delayed environments and can be seamlessly integrated with standard RL algorithms without requiring any additional modifications.
    \item \textbf{Superior performance:} Using SAC as the decision module, extensive experiments on Gym and Mujoco demonstrate that DEER achieves competitive or superior learning efficiency and performance compared with previous state-of-the-art methods.
\end{itemize}

The remainder of this paper is organized as follows. Section \ref{related_work} provides a comprehensive review of relevant techniques in reinforcement learning (RL), focusing on offline-assisted online RL and the use of encoders in RL. Section \ref{preliminary} presents essential background knowledge on Markov Decision Processes (MDP) and Random Dropping Delayed Markov Decision Processes (RDDMDP), establishing the foundation for the subsequent discussion. Section \ref{method} details our proposed algorithm, DEER. In Section \ref{Experiment}, we perform a thorough numerical validation and in-depth analysis to evaluate the performance and effectiveness of DEER across various scenarios. Finally, Section \ref{conclusion} offers valuable insights and discusses potential directions for future research in this domain.

\section{Related work}
\label{related_work}

\subsection{Offline assisted Online RL}

Numerous studies have focused on enhancing an agent's online performance with the aid of offline RL techniques. 
These studies can be categorized as follows:

(1) \textbf{Combining offline data with online learning.} Several early works attempted to initialize a replay buffer with the demonstration data \cite{1,2}. 
Other studies \cite{3,4,5,6,7} introduced new prioritized sampling schemes to improve learning efficiency and control distribution shift during the online learning stage.

(2) \textbf{Pretraining in representation or policy}. The former \cite{8} adopted standard contrastive learning methods to extract the features from a variety of offline datasets, which can be applied to downstream tasks including online RL, imitation learning and offline policy optimization. 
The latter \cite{9,10,11,12,13} , referred to as offline-to-online RL, has gained popularity in recent years and typically involves executing offline RL algorithms followed by online fine-tuning, including parameter transferring, policy regularization, etc.

Our method aligns with the key concept presented in \cite{8}, yet features significant distinctions in data source, loss function, and working principle. 
Specifically, we develop an encoder-decoder model to map information states, consisting of the delayed state and subsequent action sequence, into a common hidden space. 
This model is trained on a dataset primarily containing random data, supplemented with a minor portion of expert data from undelayed environments.

\subsection{Encoders in RL}

Encoders have gained widespread usage in RL for extracting representations as input to the policy. 
The RL4Rec framework \cite{2-rl4rec-1,2-rl4rec-2} employs a state encoder to compress users' historical interactions into a dense representation, effectively, capturing user preferences for further inference. 
Another study \cite{2-rl4rec-3} evaluated various state encoders and found that an attention-based variant can produce the optimal recommendation performance. 
Generally, encoders in RL4Rec are trained in an end-to-end manner with RL algorithms, distinguishing them from our approach.
In visual RL, pretrained encoders are employed to efficiently extract visual features and reduce image input dimensions. 
Studies such as \cite{2-visual-1} and \cite{2-visual-2} showed that pretrained ResNet representations can achieve performance comparable to state-based inputs, particularly when supplemented with expert demonstrations. 
Furthermore, \cite{2-visual-3} explored the efficacy of the image encoder in enabling agents to generalize to unseen visual scenarios with a substantial distributional shift in a zero-shot manner. 
Additionally, \cite{2-marl-1} employed a multi-view state encoder to process input states from multiple perspectives, enhancing generalization abilities via adaptive traffic signal control transfer learning. 
Despite these advances, the exploitation of pretrained models in delay scenarios remains underexplored in the current literature.

\section{Preliminary}
\label{preliminary}

\subsection{Markov Decision Process (MDP)}
\label{MDP}

The sequential decision-making problem is typically formulated as a discounted Markov Decision Process (MDP), denoted by a tuple $(\mathcal{S},\mathcal{A}, \rho,p,r,\gamma)$. Here, $\mathcal{S}$ and $\mathcal{A}$ are state and action spaces, respectively; $\rho$ is the initial state distribution; $p: \mathcal{S} \times \mathcal{A} \rightarrow \mathcal{S}$ is the transition function; $r: \mathcal{S} \times \mathcal{A} \rightarrow \mathbb{R}$ gives the reward to any transitions and $\gamma \in [0,1)$ is a discount factor. During the interaction between the agent and the environment, the agent follows a policy $\pi: \mathcal{S} \rightarrow \mathcal{A}$, resulting in a sequence of transitions or an entire trajectory $\tau=(s_t, a_t, r_t)_{t \ge 0}$. The cumulative return is calculated as $R(\tau) = \sum_{t=0}{\gamma}^{t}{r_t}$ and the primary objective in RL is to identify a return-maxmizing policy $\pi^{*}=argmax_{\pi}\mathbb{E}[R(\tau)]$.

\subsection{Random Dropping Delayed Markov Decision Process (RDDMDP)}

In real-world scenarios such as remote control and distributed communication, delays from long-distance transmission or data transfers, denoted by intrinsic delay $d_I$, significantly impact agent performance.
In addition to the always observable initial state, subsequent steps may experience state dropout during information transmission due to obstacles or network issues and the dropout probability follows a Bernoulli distribution with parameter $\mu$.
Furthermore, the continuous maximum number of extra dropping steps based on $d_I$, labeled as $d_M$, is defined to ensure that the overall delay is within the agent's capacity limit.
Therefore, at each time step $t$, the agent is expected to receive a state $s_{t-d_I}$ and a corresponding reward $r_{t-d_I}$. 
State dropout is modeled by $\omega_t \sim Bern(\mu)$, where $\omega_t = 0$ means the agent receives complete information including the state and reward, and $\omega_t = 1$ means it receives nothing.

As a result, the agent works in an environment with inherent random delays, deviating from the concept discussed in \cite{RDDP-1} and \cite{RDDP-2}. A detailed elaboration on these discrepancies is provided in Appendix \ref{appendix_discussion_1}.
The Random Dropping Delayed Markov Decision Process (RDDMDP) is proposed as follows:

\begin{definition}
\label{RDDMDP}

The RDDMDP can be defined as a 9-tuple $(d_I, d_M, \bm{\mathcal{I}_z},\bm{\mathcal{A}},\bm{\rho},\bm{p},\bm{r},\gamma,\mu)$:

(1) Intrinsic delay value: $d_I \in \mathbb{Z}^+$, which is caused by long distance transmission or heavy data transfers;

(2) Continuous maximum number of extra dropping steps: $d_M \in \mathbb{N}$, which is defined to ensure that $d_I+d_M$ remains within the agent's capacity;

(3) Information state space:  $\bm{\mathcal{I}_z}=\mathcal{S}\times\mathcal{A}^z$, where $z$ denotes the random delay value with $d_I \le z \le {d_I + d_M}$, $\mathcal{S}$ and $\mathcal{A}$ are the same as the definition in MDP ;

(4) Action space:  $\bm{\mathcal{A}}=\mathcal{A}$;

(5) Initial information state distribution: 

$\bm{\rho}(\bm{i}_0)=\bm{\rho}(s_0,a_0,...,a_{d_{I}-1})=\rho(s_0)\prod_{i=0}^{d_I-1}\delta(a_i-c_i)$, 

where $\rho$ is the initial state distribution in MDP and $\{c_i\}_{i=0}^{d_I-1}$ are actions selected randomly at the initial of trajectories when states are not observed, and  $\delta$ is the Dirac delta function;

(6) Transition distribution:  $\bm{p}(\bm{i}_{t+1}|\bm{i}_{t},\bm{a}_t)$, where $\bm{a}_t \in \bm{\mathcal{A}}$ and the information state $\bm{i}_{t} \in  \bm{\mathcal{I}_z}$ is described in detail below;

(7) Reward function: $\bm{r}_t=r_{t-z_t}$, where $z_t$ denotes the random delay value at time $t$;

(8) Discount factor: $\gamma \in [0,1)$;

(9) Dropping probability: $\mu\in[0,1)$, and when $\mu=0$, the RDDMDP is reduced to the constant delayed MDP (CDMDP) and the details are provided in Appendix \ref{CDMDP}.

\end{definition}
At each time $t$, there is a chance of $\mu$ that the agent does not receive the delayed state $s_{t-d_I}$, leading to a potential dropout of state. Thus, the random delay value $z_t$ is defined in the following manner:
\begin{equation*}
    z_{t}= \begin{cases} d_I,  & \text{with probability } 1 - \mu ,\\ 
    {z_{t-1} + 1}, & \text{with probability $\mu$ and $z_{t-1} < d_I+d_M$}, \\ 
    {d_I+d_M}, & \text{others.} \end{cases}
\end{equation*}
The information state $\bm{i}_{t}$ is defined correspondingly as 
\footnote{
The superscript of $a_{t_1}^{(t_2)}$ shows that the action is an element of the information state $\bm{i}_{t_2}$ and the subscript indicates that the action is taken at timestep $t_1$.}
:

 \textbullet \ \ $ \bm{i}_{t} = (s_{t-z_t}, (a_{t-n}^{(t)})_{n=z_t:1}))$, with probability  1 - $\mu$;

 \textbullet \ \ $ \bm{i}_{t} = {concatenate(\bm{i}_{t-1}, a_{t-1})}$, with probability $\mu$ and $z_{t-1} < d_I+d_M$;

 \textbullet \ \ $ \bm{i}_{t} = {concatenate(\bm{i}_{t-1}[:dimenson(\mathcal{S})], (a_{t-n}^{(t)})_{n=z_t:1}))}$, others.

Accordingly,  the reward function is expressed as:
\begin{equation*}
  \begin{aligned}
    \bm{r}_{t} =    
    \begin{cases} r_{t-d_I},  & \text{with probability } 1 - \mu ,\\ 
    {\bm{r}_{t-1} }, & \text{others.} 
    \end{cases}
  \end{aligned}
\end{equation*}

After modeling the delays as mentioned earlier, the agent will proceed to take actions based on its current information state $\bm{i}_t$.
A concise example is provided in Appendix \ref{demonstration}.

\section{Method}
\label{method}

In this section, we present Delay-resilient Encoder-Enhanced RL (DEER), a succinct and effective framework designed to tackle delays in RL. 
DEER leverages an encoder pretrained on undelayed datasets to extract informative features, enabling it to adeptly manage both constant and random delays. 
The algorithmic framework of DEER is detailed in Algorithm \ref{algorithm}.

\subsection{Pretrained Encoder}
\begin{figure*}[t]
    \centering
    \includegraphics[width=0.85\textwidth]{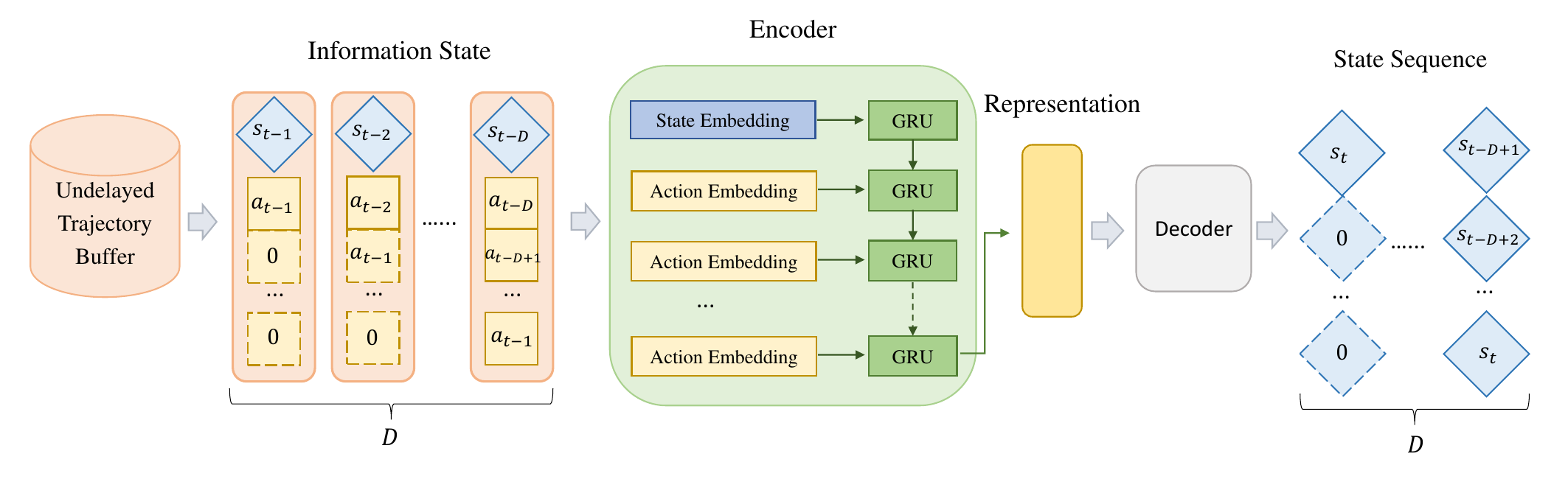}
    \caption{Process of model pretraining. Firstly, the information state dataset is created based on the original undelayed dataset. All state sequences are standardized to a uniform length $D$, where $D$ represents the maximum delay in the environment. Next, these datasets are fed into the Seq2Seq model and trained in a supervised manner.}
    \label{fig2}
\end{figure*}

DEER effectively leverages pretrained models as feature extractors, requiring no modification of the RL algorithm. 
The pretrained encoder projects information states into embeddings of uniform length, facilitating the agent's management of delay challenges without the prior knowledge of environment delays. 
Throughout the policy learning process across training tasks, the encoder's parameters remain fixed to acquire universal context representations.
To ensure the proficiency of the encoder, the training of the encoder-decoder model is conducted on the datasets composed of trajectories generated by a random policy along with a few expert trajectories collected by a well-trained SAC agent, all sourced from undelayed environments.
The input and output of the model are denoted as the information state $I_t = (s_t,a_{t},...,a_{t+d-1})$ and the state sequence $(s_{t+1},...,s_{t+d})$, respectively. 
The encoder-decoder model operates as a regression model for state sequence prediction. 
Given the capabilities of the encoder-decoder model, the hidden features extracted by the encoder are expected to contain valuable insights into delays, thereby aiding the agent in making informed decisions.
Furthermore, to facilitate the encoder's generalization across various constant delays and effectively handle random delays, the training dataset consists of information states with diverse action sequence lengths, while maintaining a consistent dimensionality for the hidden features. 
This approach ensures that the encoder can acquire features directly applicable to the agent, irrespective of the specific delay conditions.

We adopt a Seq2Seq \cite{seq2seq} model as the encoder-decoder framework, a succint yet effective approach for handling the delay issue. 
Initially, Multi-Layer Perceptrons (MLPs) are applied to encode each component of the information state, including a state and a series of actions, to generate corresponding embeddings. 
These embeddings are then inputted into a GRU module to produce the hidden feature vector whose dimension is a hyperparameter.
The Seq2Seq model is trained using the MSE loss to enhance the accuracy of state sequence predictions, thereby refining the representation of the information state's hidden features. 
The complete process of model pretraining is illustrated in Fig.\ref{fig2}, and the detailed network structure and parameter configurations are provided in Appendix \ref{appendix_implementaion_encoder}.

\subsection{Encoder-enhanced Policy Learning}

The pretrained encoder assumes a pivotal role during the policy learning phase by extracting essential representations of delayed information, thereby enabling standard RL algorithms to learn effectively, regardless of environment delays.
By providing the context representation based on delayed information, the encoder offers distinct advantages across both constant and random delay environments. 
In constant delay settings, its efficacy lies in its ability to generalize across diverse delay types, owing to its training on universal datasets. 
This facilitates the direct transformation of information states with unknown lengths into fixed-length representations, obviating the need for adjustments to policy input dimensions. 
In random delay environments, original information states of varying lengths are encoded into hidden features of constant lengths, facilitating the seamless integration of standard RL algorithms that rely on fixed-length inputs. 
The entire process of the encoder-enhanced policy learning is shown in Fig.\ref{fig1}.

\section{Experimental results}
\label{Experiment}
In this section, we systematically evaluate the effectiveness of our approach by comparing it with state-of-the-art RL algorithms in both constant and random delay environments. 
We analyze the impact of the dimension of context representation and the composition of the pre-training dataset on the final task performance. Additionally, ablation experiments are conducted to elucidate the rationale behind the design of the DEER structure. 
Finally, we conduct a time comparison of different algorithms and discuss the limitations of DEER.

\begin{figure*}[t]
\centering
\includegraphics[width=0.75\textwidth]{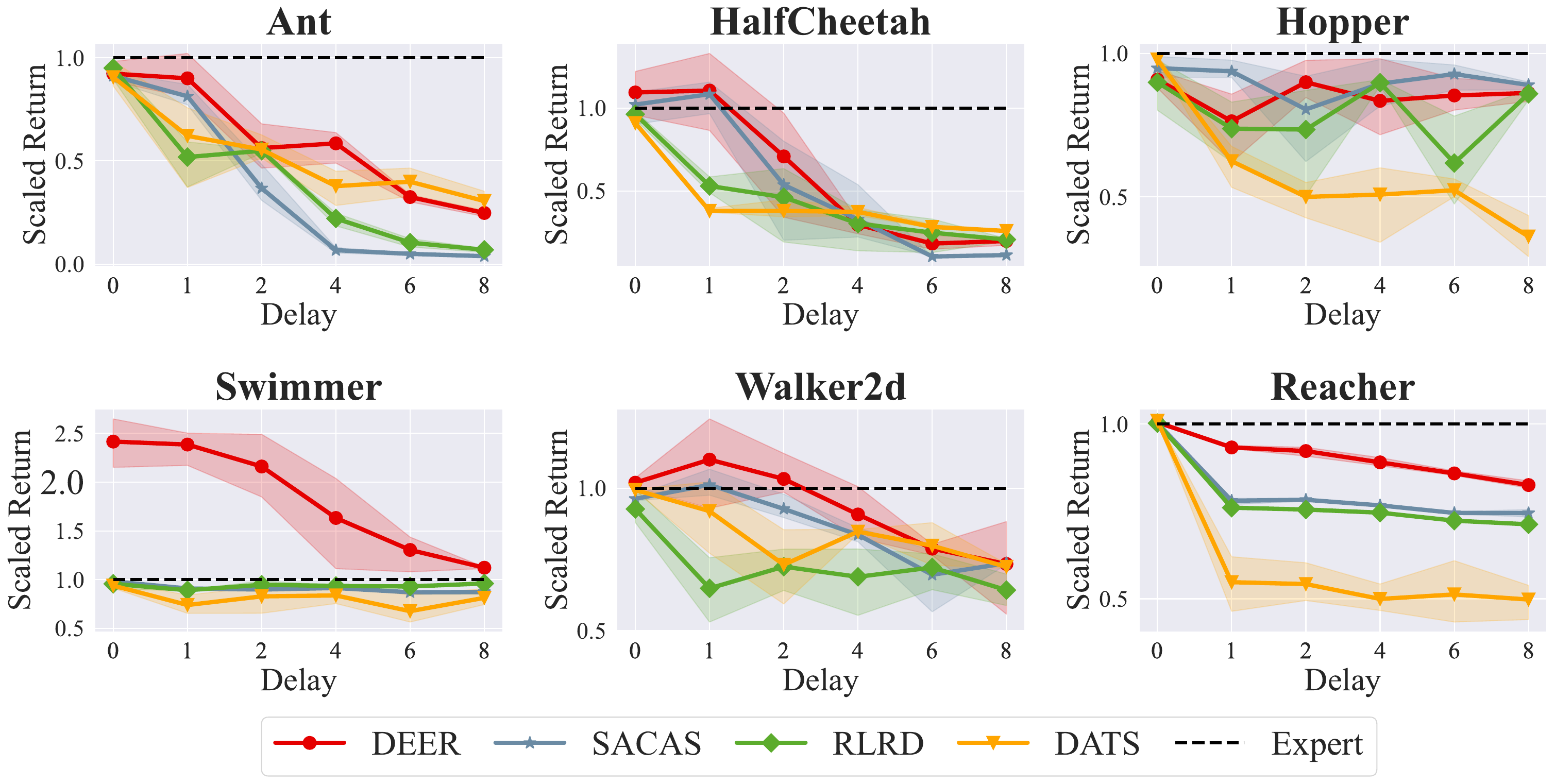}
\centering
\caption{Comparison of algorithms under diverse constant delays.}
\label{comparision with constant delay}
\end{figure*}

We use SAC for decision making, a popular choice for continuous control tasks due to its integration of the actor-critic architecture and the maximum entropy principle. 
When the context representation is generated by the pretrained encoder, the agent takes action based on the new state and updates its policy, similar to its behavior in undelayed environments.

All experiments are conducted under the MuJoCo environments from the gym library, including Ant, HalfCheetah, Hopper, Swimmer, Walker2d, and Reacher. Each algorithm is executed with 5 different seeds in each environment. 
The shaded regions in all figures represent 95\% confidence intervals, while the ``+/-" uncertainties in all tables represent variances.
The details regarding the number of trajectories used in the pretraining phase are provided in Appendix \ref{offline datasets}.

\subsection{Evaluation}
\label{sec:evaluation}
The following algorithms are used in comparative studies to illustrate the effectiveness of our proposed method:
\begin{itemize}
    \item \textbf{Reinforcement Learning with Random Delays (RLRD \cite{2-modelfree-7}).} RLRD introduces a technique where past actions are relabeled using the current policy. This relabeling procedure generates on-policy sub-trajectories, providing an off-policy and planning-free approach applicable to environments with constant or random delays.
    \item \textbf{Delay-Aware Trajectory Sampling (DATS \cite{2-modelbased-3}). } The effectiveness of DATS can be attributed to the synergistic combination of its unique dynamics model and its effective planning method, PETS.
    The dynamics model incorporates both the known component resulting from delays and the unknown component inherited from the original MDP.
    \item \textbf{Soft Actor-Critic with Augmented States (SACAS).} The implementation of SACAS aligns with the principles described in \cite{2-modelfree-1}.
\end{itemize}

Considering the differences in reward settings between DATS and other methods, we normalize the cumulative rewards by $\frac{\bm{Return - min\_return}}{\bm{Expert\_return - min\_return}}$. 
The parameters remain consistent within each algorithm but may vary across different algorithms. 
$\bm{Return}$ represents the cumulative rewards obtained in each episode; $\bm{min\_return}$ corresponds to the minimum return observed throughout all experiments; 
$\bm{Expert\_return}$ indicates the level of expertise achieved in undelayed environments.

\textbf{Constant Delays.} The initial experiments focus on environments with constant delays. Four algorithms are compared in environments where delay values are set to 0, 1, 2, 4, 6 and 8, respectively.
The environment with a delay value of 0 signifies a delay-free setting, serving as a benchmark for evaluating the baseline performance of the compared algorithms.

\begin{figure*}[t]
\centering
\includegraphics[width=0.75\textwidth]{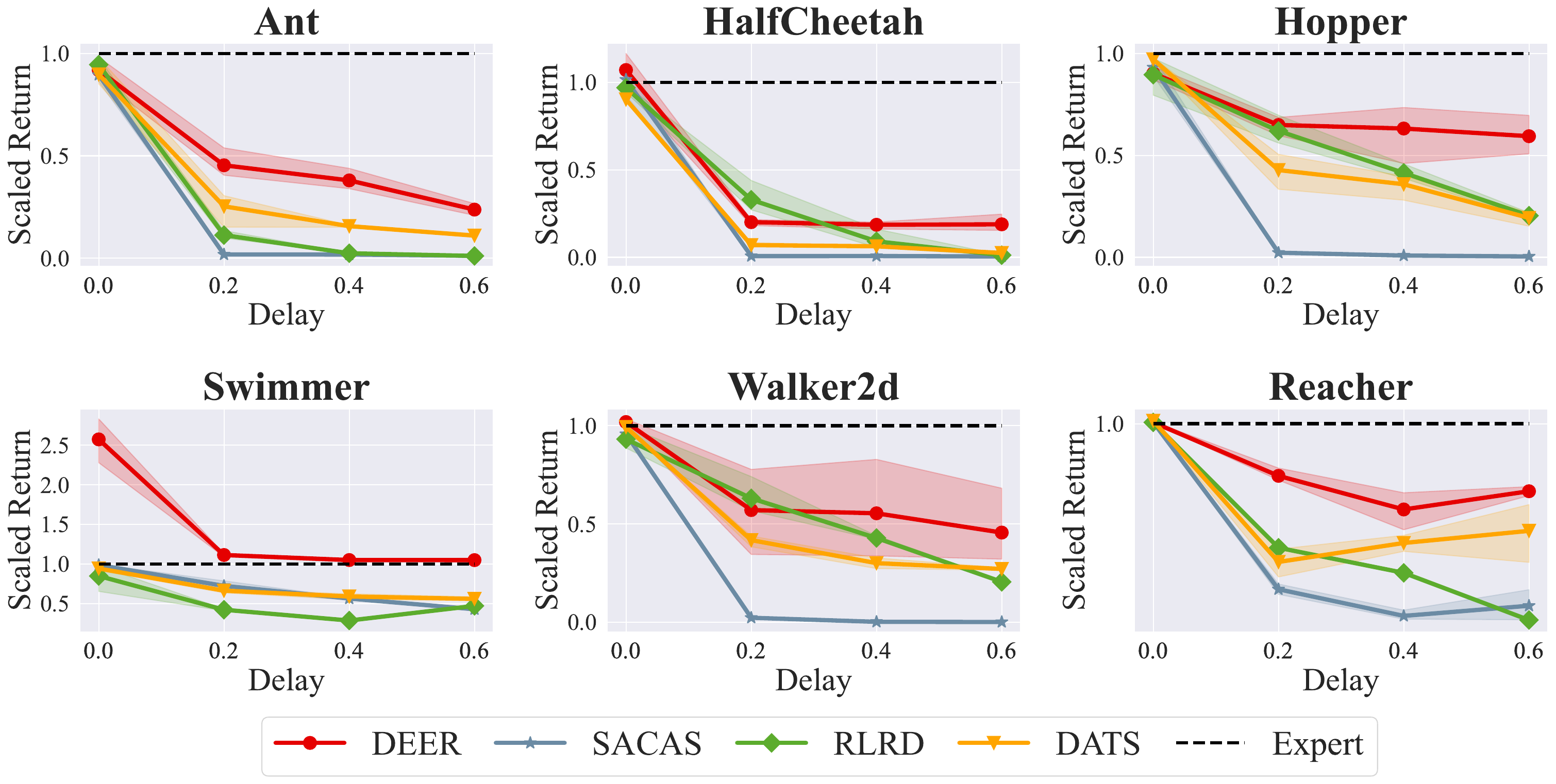}
\centering
\caption{Comparison of algorithms under diverse random delays}
\label{comparision with random delay}
\end{figure*}

\begin{table*}[t]
  \centering
  \caption{Comparison of DEER’s performance in various dimensions with different delay values or dropping probabilities.}
    \renewcommand\arraystretch{1.2}
    \begin{tabular}{cc|ccccc|ccc}
    \toprule
    \multicolumn{2}{c|}{\multirow{2}{*}{\bf Setting}} & \multicolumn{5}{c|}{\bf Delay} & \multicolumn{3}{c}{\bf Drop} \\
    \cmidrule(lr){3-7}\cmidrule(lr){8-10}
    &  & \bf 1     &  \bf 2  & \bf 4  & \bf 6  & \bf 8 & \bf 0.2     & \bf 0.4  & \bf 0.6 \\
    \midrule
    \multicolumn{1}{c|}{\multirow{3}{*}{\bf Dimension}}
    & \bf 128 & 1.07±0.35 & 0.91±0.37  & 0.69±0.36 & 0.61±0.39  & 0.51±0.37 & 0.67±0.34  & 0.64±0.33 & \bf 0.62±0.34 \\
    \multicolumn{1}{c|}{}& \bf 256 & 1.21±0.48  & \bf 1.05±0.47  & \bf 0.86±0.46  & \bf 0.72±0.41 & \bf 0.66±0.37  & 0.76±0.47 & \bf 0.66±0.31 & \bf 0.62±0.34\\
    \multicolumn{1}{c|}{} & \bf 512 & \bf 1.22±0.51  & 0.95±0.45   & 0.78±0.51  & 0.72±0.45 & 0.60±0.38  & \bf 0.78±0.62 & 0.65±0.50  & 0.55±0.39\\
    \bottomrule
    \end{tabular}%
  \label{tab:various dimensions with different delays or dropping probabilities}%
\end{table*}%

As shown in Figure \ref{comparision with constant delay}, it is clear that: 
1) All compared algorithms can attain or surpass expert-level performance in tasks without delays. However, as the delay in the environment increases, their performance gradually diminishes; 
2) In Ant, Swimmer, Walker2d, and Reacher, DEER outperforms other algorithms, as evident from their respective performance curves, while in HalfCheetah and Hopper, DEER's performance is similar to that of other algorithms or slightly lower with certain delay values (e.g., Hopper with a delay of 4); 
3) DEER consistently outperforms the expert in Swimmer across various delays, further highlighting the effectiveness of the context representation in making informed decisions. A more detailed analysis of DEER's remarkable results in Swimmer is offered in Appendix \ref{Swimmer Analysis}.

\textbf{Random Delays.} Randomly delayed environments pose a tougher challenge compared with constant delays due to the increased risk of information dropout. 
We evaluate the four aforementioned algorithms with $d_I=2$, $d_M=4$, and dropping probabilities $\mu=0.2, 0.4, \text{and } 0.6$, respectively.

From Figure \ref{comparision with random delay}, the following observations can be made:
1) Across varying dropping probabilities, DEER consistently outperforms other methods in the Ant, Swimmer, and Reacher environments. 
Particularly noteworthy is the substantial superiority of DEER over the expert level in the Swimmer environment. 
2) In the HalfCheetah environment, DEER ranks second only to RLRD at a dropping probability of 0.2, and as the dropout rate increases, DEER demonstrates a slight performance advantage over other methods. 
3) In the Hopper and Walker2d environments, DEER's performance is comparable to other methods at a dropping probability of 0.2, while in the other two settings, DEER exhibits superior efficacy. 

In summary, the context representation generated by DEER's pretrained encoder can effectively extract valuable information from delayed states, readily applicable across varying delay settings.

\subsection{Impact of Key Factors}

The context representation plays a crucial role both as input to the decision model and within the encoder-decoder architecture.
In this section, we delve into the two most important factors related to DEER: the dimension of context representation and the composition of the dataset used to train the encoder.

\textbf{Dimension.} 
Table \ref{tab:various dimensions with different delays or dropping probabilities} illustrates the average performance of different dimensions of context representation across various environmental settings (delays or dropping probabilities) for the six tasks. 
It reveals  that the agent can effectively make decisions based on context representations of different dimensions, with notably superior performance when the dimension is set to 256. 
The high data variance in the table is attributed to the agent consistently outperforming the expert in the Swimmer task under varying environmental conditions. 
This observation indicates that, despite a larger dimension of representation, better overall performance for the agent is not guaranteed with a consistent training strategy.

The final performance of the agent relies on the representation capability of the pretrained encoder. 
It is advantageous for decision-making only when the context representation adequately captures delayed information, indicating the pretrained encoder's proficiency in representing the information states. 
Therefore, the dimension of representation does not necessarily correlate positively with the final performance.
In conclusion, considering factors such as computational complexity and overall performance, opting for a 256-dimensional context representation is generally recommended.

\textbf{Composition.}
The existing dataset, predominantly comprised of random trajectories with a small fraction of expert trajectories, is designed to enhance realism in mimicking real-world scenarios. 
To evaluate the impact of offline datasets on the model, we introduced two supplementary datasets for these six tasks with a delay of 4: 
one comprised entirely of random trajectories and the other consisting solely of expert trajectories. 
The number of trajectories used in these datasets are provided in Table \ref{tab: offline dataset}.
The training process for the two new datasets remain consistent with DEER's, and the comparative results are presented in Table \ref{tab:analysis about offline datasets}.

As shown in the table, when the dataset comprises exclusively random or expert trajectories, the performance notably legs behind that of the current mixed dataset, except in the Hopper environment. 
This outcome underscores the significant influence of encoder quality on the final performance of the agent. 
Furthermore, it can be inferred that a substantial number of random trajectories can capture a broader range of initial conditions, whereas expert trajectories can provide more extensive trajectory segments.
This combination yields a more comprehensive dataset for encoder training, thereby facilitating more informed decision-making by the agent.

\begin{table*}[htbp]
  \centering
  \caption{Comparison of different offline datasets}
  \renewcommand\arraystretch{1.2}
    \begin{tabular}{c|c|ccc}
    \toprule
    \multicolumn{2}{c|}{\bf Offline datasets } & \bf Random & \bf Large random and few expert & \bf Expert \\
    \midrule
    \multirow{6}{*}{\bf Tasks} 
    & \bf Ant & 869.75 $\pm$ 80.53 & \bf 2987.07 $\pm$ 1421.40 & 863.05$\pm$98.97 \\
    & \bf HalfCheetah & 987.14$\pm$317.79 & \bf 5780.5$\pm$543.8 & 230.91$\pm$282.98 \\
    & \bf Hopper & 2972.85$\pm$830.32 & 2918.8$\pm$556.4 & \bf 3061.22$\pm$640.34 \\
    & \bf Swimmer & 36.64$\pm$2.31 & \bf 84.85$\pm$28.06 & 33.15$\pm$2.96 \\
    & \bf Walker2d & 415.24$\pm$128.33 & \bf 4119$\pm$430.8 & 509.50$\pm$99.23 \\
   & \bf Reacher & -18.49$\pm$3.40 & \bf -7.96$\pm$2.07 & -18.80$\pm$3.09 \\
    \bottomrule
    \end{tabular}%
  \label{tab:analysis about offline datasets}%
\end{table*}%

\begin{table*}[t]
  \centering
  \caption{Comparison of DEER and DOLPS under delay values of 4 and 6}
  \renewcommand\arraystretch{1.2}
    \begin{tabular}{c|cc|cc}
    \toprule
    \bf Delay & \multicolumn{2}{c|}{\bf 4} & \multicolumn{2}{c}{\bf 6} \\
    \midrule
    \bf Algorithm & \bf DEER  & \bf DOLPS & \bf DEER  & \bf DOLPS \\
    \midrule
    \bf Ant   & \bf 2987.07$\pm$1421.40  & -442.68$\pm$643.98 & \bf 1639.98$\pm$485.07  & -38.34$\pm$30.61 \\
    \bf HalfCheetah & \bf 5780.5$\pm$543.8  & 2702.77$\pm$480.21  & \bf 3853.7$\pm$145.3  & 1743.20$\pm$436.93 \\
    \bf Hopper & \bf 2918.8$\pm$556.4  & 825.54$\pm$495.56   & \bf 3017.25$\pm$674.95  & 335.57$\pm$11.17 \\
    \bf Swimmer & \bf 84.85$\pm$28.05    & 41.80$\pm$1.36    & \bf 88.84$\pm$24.27   & 37.76$\pm$12.40 \\
    \bf Walker2d & \bf 4119$\pm$430.8  & 381.53$\pm$113.33   & \bf 3546$\pm$1100.4  & 234.68$\pm$93.61 \\
    \bf Reacher & \bf -7.96$\pm$2.07  & -17.36$\pm$3.25   & \bf -9.65$\pm$2.00  & -20.98$\pm$4.16 \\
    \bottomrule
    \end{tabular}%
  \label{tab:ablation DOLPS}%
\end{table*}%

\subsection{Ablation Study}

In this subsection, we will address the following four questions: 
1) What would be the impact if we bypass context representation and make decisions directly based on the states predicted by the decoder? 
2) How would the performance be affected if we opt for online training of the encoder without pretraining it on offline datasets? 
3) If we apply the same dataset as DEER and employ state-of-the-art offline-to-online method, what outcomes would occur?
4) When the offline trajectories trained by the encoder used for DEER are applied to the algorithms mentioned in Section \ref{sec:evaluation}, does it enhance their performance?

\textbf{Question 1.}
We introduce a comparative method called DOLPS, which stands for Decision on Last Predicted State. 
Delayed states can be inferred from the decoder module trained with the encoder in DEER simultaneously.
Experimental results in Table \ref{tab:ablation DOLPS} consistently demonstrate DEER's advantage over DOLPS across various environments and delays, with DOLPS showing limited effectiveness in Ant, Hopper, and Walker2d.
It is evident that context representation not only effectively mitigates prediction errors crucial for decisions in the original decision space but also captures historical information embedded in delayed states and action sequences, providing an advantage for decision-making in delayed scenarios.

\textbf{Question 2.} 
The comparative method, referred to as online-DEER, shares the same structure as DEER, including an encoder-decoder and a decision-making component. 
Initially, its encoder-decoder network is randomly initialized, and training data is collected through real-time interactions within the environment. 
The network is updated every 300,000 steps. 
All hyperparameters and training process are consistent with those of DEER.

Comparative experiments are conducted on six distinct tasks, each configured with a delay set at 4.
The corresponding results are shown in Table \ref{tab:comprehensive online deer}. 
As depicted in the table, the performance of DEER notably surpasses that of online-DEER.
This disparity can be attributed to the continuous variation of the encoder during the training process of online-DEER, leading to fluctuations in the context representation even for the same augmented state, 
which further impacts the decision-making of the agent. 
It underscores the critical role of stability and accuracy in context representation for effective decision-making and highlights the importance of pretraining.

\begin{table}[t]
  \centering
  \caption{Comparison of Online DEER and Offline DEER with a delay of 4 }
  \renewcommand\arraystretch{1.2}
    \begin{tabular}{ccc}
    \toprule
   \bf Version of Deer & \bf Online & \bf Offline \\
    \midrule
    \bf Ant & 883.43$\pm$70.1 & \bf 2987.07$\pm$1421.40 \\
    \bf HalfCheetah & 1590.14$\pm$1453.23 & \bf 5780.5$\pm$543.8 \\
    \bf Hopper & 736.50$\pm$585.83 & \bf 2918.8$\pm$556.4 \\
    \bf Swimmer & 44.57$\pm$1.76 & \bf 84.85$\pm$28.05 \\
    \bf Walker2d & 144.88$\pm$180.35 & \bf 4119$\pm$430.8 \\
    \bf Reacher & -11.26$\pm$3.16 & \bf -7.96$\pm$2.07 \\
    \bottomrule
    \end{tabular}%
  \label{tab:comprehensive online deer}%
\end{table}%

\textbf{Question 3.}
The selected state-of-the-art offline-to-online algorithm for comparison is PEX \cite{zhang2023policy}.
PEX's core approach involves initially learning a policy from an offline dataset and using this learned policy as a candidate for further learning by another policy, with both policies interacting with the environment in an adaptable manner. 
During the training of the offline policy, PEX requires reward information in addition to states and actions to guide its learning, whereas DEER relies solely on states and actions.

We conduct policy training using PEX across three continuous tasks (HalCheetah, Hopper, and Walker2d), incorporating delays of 4, 6, and 8. 
Each scenario is experimented on using three different seeds. 
The offline policy training dataset for PEX aligns with that of DEER.
The comparative experiment results are displayed in Table \ref{tab:SOTA}. 
Analysis of the data in the table indicates that the final training outcome of PEX closely resembles that of an agent employing a random strategy. 
The subpar performance is likely associated with the utilized offline dataset, primarily constituted by a significant majority of random trajectories alongside a limited number of expert trajectories. 
This composition within the offline data has led PEX to learn a strategy leaning towards randomness, consequently impacting the agent's online decision-making. 
In contrast, DEER, utilizing the same dataset, learns a representation of the information state that enhances the agent's decision-making during the online process.

\begin{table*}[t]
  \centering
  \caption{Comparison of PEX and DEER}
  \renewcommand\arraystretch{1.2}
    \begin{tabular}{c|cc|cc|cc}
    \toprule
    \bf Delay  & \multicolumn{2}{c|}{\bf 4} & \multicolumn{2}{c|}{\bf 6} & \multicolumn{2}{c}{\bf 8} \\
    \midrule
   \bf Algorithm & \bf DEER  & \bf PEX   & \bf DEER  & \bf PEX   & \bf DEER  & \bf PEX \\
    \midrule
   \bf  HalfCheetah  &  \bf 5780.5$\pm$543.8   & 1038.1$\pm$289.3	 & \bf 3853.7$\pm$145.3   & 957.0$\pm$357.5   & \bf 2924.7$\pm$456.7 & 859.6$\pm$298.8 \\
    \bf Hopper & \bf 2918.8$\pm$556.4   & 7.7$\pm$0.2   & \bf 3017.25$\pm$674.95  & 6.5$\pm$0.1   & \bf 2462$\pm$335.5  & 7.1$\pm$0.2 \\
    \bf Walker2d  & \bf 4119$\pm$430.8  & 1.6$\pm$0.2   & \bf 3546$\pm$1100.4  & -4.8$\pm$0.4  & \bf 3074$\pm$1308.2  & -4.5$\pm$0.7 \\
    \bottomrule
    \end{tabular}%
  \label{tab:SOTA}%
\end{table*}%

\begin{table*}[t]
  \centering
  \caption{Comparison of algorithms with DEER's offline datasets }
  \renewcommand\arraystretch{1.2}
    \begin{tabular}{c|cc|cc|cc}
    \toprule
    \bf Env  & \multicolumn{2}{c|}{\bf HalfCheetah} & \multicolumn{2}{c|}{\bf Hopper} & \multicolumn{2}{c}{\bf Walker2d} \\
    \midrule
   \bf Case & \bf before  & \bf after   & \bf before  & \bf after   & \bf before  & \bf after \\
    \midrule
    \bf  DEER  &  0.28$\pm$0.01   &  \bf{0.28$\pm$0.01} \textbf{-}	 & \bf{0.75$\pm$0.08}   & \bf{0.75$\pm$0.08}  \textbf{-} & \bf{0.80$\pm$0.05} & \bf{0.80$\pm$0.05} \textbf{-} \\
    \bf SACAS & 0.27$\pm$0.01  & 0.0$\pm$0.1  $\downarrow$  & 0.78$\pm$0.02  & 0.0$\pm$0.1  $\downarrow$ & 0.78$\pm$0.01  & 0.0$\pm$0.1 $\downarrow$ \\
    \bf RLRD  & 0.28$\pm$0.01  & 0.26$\pm$0.1 $\downarrow$  & \bf{0.78$\pm$0.01}  & 0.51$\pm$0.01 $\downarrow$ & 0.62$\pm$0.13  & 0.73$\pm$0.01 $\uparrow$ \\
    \bf DATS  & \bf{0.30$\pm$0.00}  & 0.18$\pm$0.01 $\downarrow$ & 0.50$\pm$0.15  & 0.46$\pm$0.01 $\downarrow$ & 0.78$\pm$0.01  & 0.48$\pm$0.03 $\downarrow$ \\
    \bottomrule
    \end{tabular}%
  \label{tab:fairness}%
\end{table*}%

\textbf{Question 4.} 
The dataset used by DEER's encoder originates from a non-delay environment , primarily comprising random trajectories, with only a minimal portion of expert trajectories. 
We integrate these datasets into the aforementioned delayed-RL algorithms as follows: 
1) For SACAS and RLRD, the datasets are structured into states, actions, and rewards following the definitions in RDDMDP and \cite{2-modelfree-7}, respectively, and are stored in their replay buffers. 
When the stored data exceeds the buffer's maximum capacity, old data is discarded.
2) DATS, being a model-based algorithm, is updated using the most recently saved trajectories. 
Before executing the DATS algorithm, we update the model with these datasets.
It is worth noting that, DEER requires only the states and actions of trajectories, without the need for reward information.

Experiments are conducted on the tasks of HalfCheetah, Hopper, and Walker2d, with delays set at 4 and the results are shown in Table \ref{tab:fairness}.
The results clearly indicate that a performance decline for all algorithms, except DEER. 
Specially, SACAS notably dropping to zero directly, while RLRD remains relatively stable in Walker2d, albeit still underperforming compared to DEER.
This performance decline can be attributed to the composition of the dataset. 
The prevalence of random trajectories directly affects the composition of the replay buffer in the original algorithms, subsequently impacting the training of the policy network. 
Over extended training periods, the model updates its parameters based on the random data, resulting in diminished agent intelligence and diminished agent intelligence its decision-making process.

\subsection{Analysis of time and limitation}

The experimental results confirm DEER's efficacy in addressing delay problems, particularly highlighting the significant performance gains achieved by well-pretrained encoders. 

\begin{figure}[t]
\includegraphics[width=0.45\textwidth]{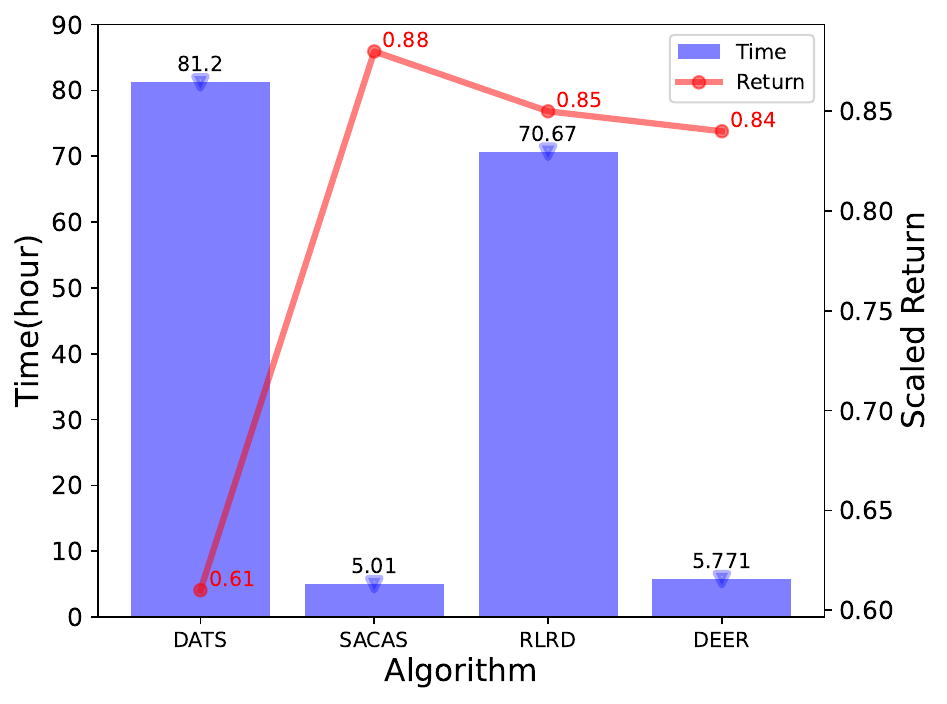}
\centering
\caption{Comparison of DEER's performance with different number of expert trajectories in Walker2d.}
 \label{pic:time}
\end{figure}

\textbf{Time Analysis.}
We conduct a comparative analysis of the time and performance of four algorithms: DEER, RLRD, DATS, and SACAS. 
In the Hopper task, a fixed delay of 4 is established, with each algorithm undergoing three runs of 1 million environmental steps using different seeds. 
All algorithms are executed on the same NVIDIA GeForce RTX 4090 graphics card. 
The time and performance comparison are illustrated in Figure \ref{pic:time}. 
 
In comparison to DEER and SACAS, RLRD and DATS incur excessive time consumption, although RLRD displays decent final performance. 
DEER requires more time than SACAS, which can be attributed to both the pretraining duration and the encoding of information states into context representation. 
While DEER performs slightly lower than SACAS in the current settings, Figure \ref{comparision with constant delay} demonstrates that DEER surpasses SACAS, particularly in the Swimmer task.

\textbf{Limitation.}
The encoder in DEER is crucial for extracting context representations from delayed information, enabling the agent to effectively handle both constant and random delays. This significantly impacts the agent's performance.
The differences in the number of trajectories used for pretraining the encoder, as outlined in Section ``Mimicking Reality" in Table \ref{tab: offline dataset}, are due to the following reasons:
1) The maximum number of steps per episode varies across different tasks. The number of trajectories is designed to ensure that each task's training dataset contains a sufficient number of state transitions, comparable to the amount needed to train an agent to expert-level performance in a delay-free environment. 
2) Trajectories generated by the random strategy primarily capture the initial state distribution and partial state transition functions of the environment, whereas expert trajectories capture the opposite characteristics. It is essential to balance the quantity of these two types of trajectories.

\begin{figure}[t]
\centering  
\subfloat[Different constant delays]{   
\begin{minipage}{8cm}
\centering    
\includegraphics[width=1\textwidth]{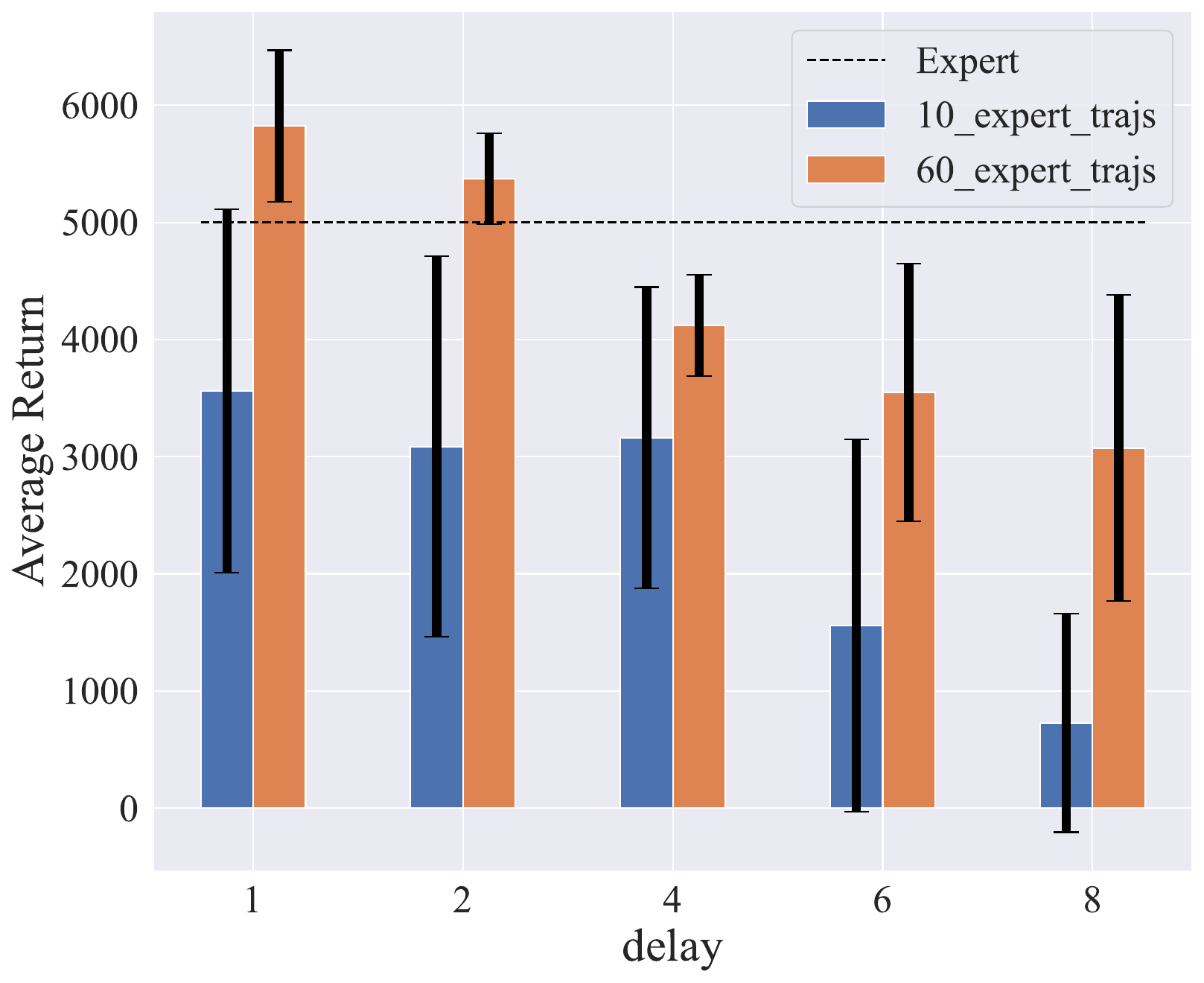} 
\end{minipage}
}

\subfloat[Different dropping probabilitie]{ 
\begin{minipage}{8cm}
\centering    
\includegraphics[width=1\textwidth]{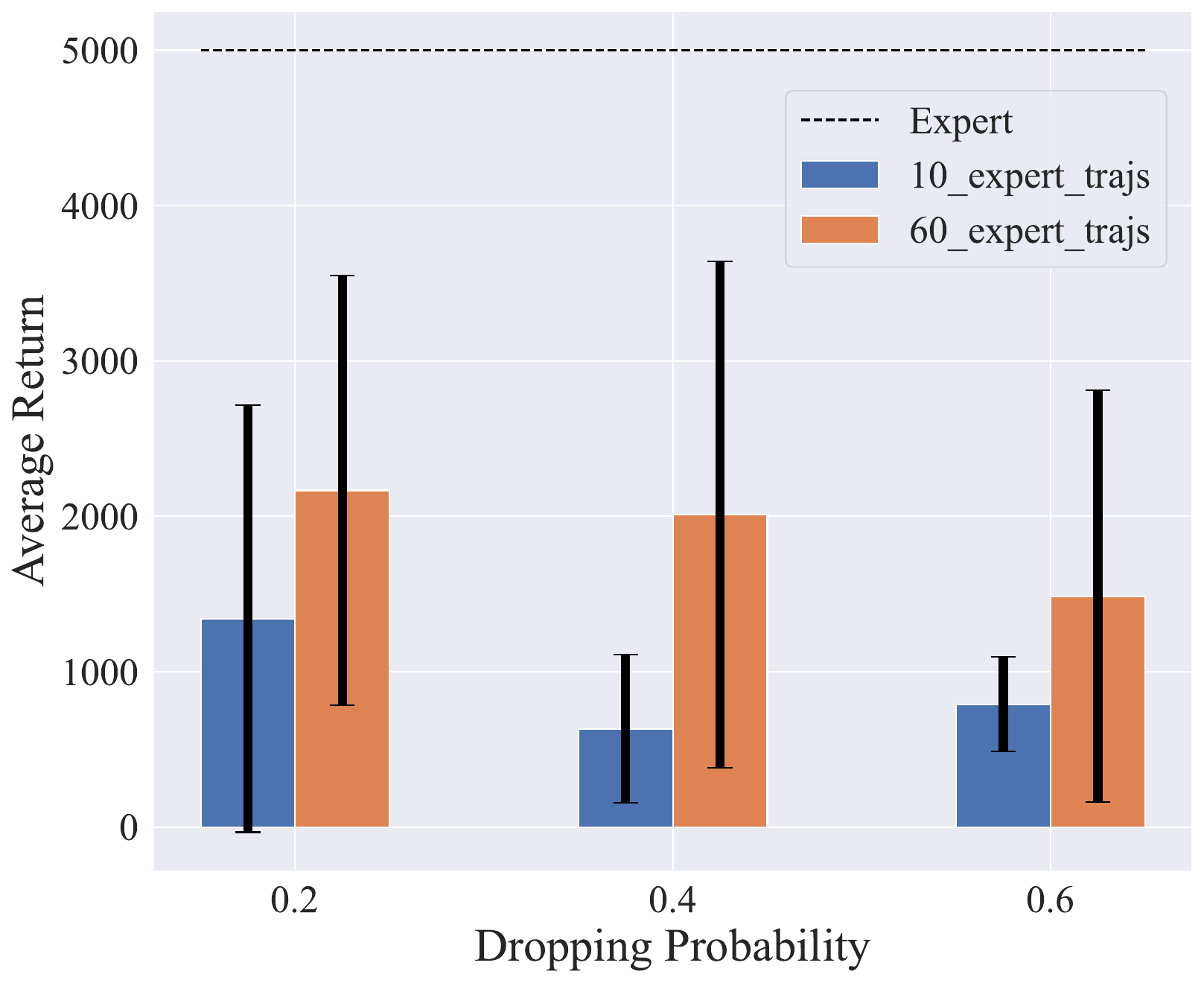}
\end{minipage}
}
\caption{Comparison of DEER's performance with different number of expert trajectories in Walker2d.}    
\label{fig: limitation}    
\end{figure}

As shown in Table \ref{tab: offline dataset}, DEER achieves satisfactory performance with only 10 expert trajectories in all tasks except Walker2d. To validate this claim, we conduct experiments by keeping the number of random trajectories constant while varying the expert trajectories to 10 and 60. Figure \ref{fig: limitation} clearly shows that increasing the number of expert trajectories improves the agent's final performance while maintaining the same number of random trajectories. Therefore, when using DEER to address new tasks with delays, it is advisable to provide the encoder with a state transition dataset that is comparable to the training data used for achieving expert performance in a delay-free environment. Additionally, it is beneficial to include as many expert trajectories as possible  to enhance the agent's performance.

\section{Conclusion and Future work}
\label{conclusion}
In this paper, we introduce DEER, a concise framework designed to effectively tackle delay issues in RL, including both constant delays and random delays, while enhancing the interpretability of the entire process. 
In DEER, an encoder is pretrained using trajectories collected from delay-free environments to map information states containing the delayed information into hidden features called context representations, which are subsequently used by the agent to derive new actions. 
Experiments on DEER combined with SAC demonstrate that our method achieves competitive or superior learning efficiency and performance in comparison with state-of-the-art methods, which validate the effectiveness and efficacy of our approach in addressing delay-related challenges. 

Future work will focus on extending DEER to visual reinforcement learning, where agents receive and process visual information as states. Additionally, efforts will be made to deploy our approach to real-world systems, such as remote control systems or physical robots, further assessing its performance and applicability in practical scenarios.


\bibliographystyle{IEEEtran}
\bibliography{deer}

\begin{thebibliography}{10}
\providecommand{\url}[1]{#1}
\csname url@samestyle\endcsname
\providecommand{\newblock}{\relax}
\providecommand{\bibinfo}[2]{#2}
\providecommand{\BIBentrySTDinterwordspacing}{\spaceskip=0pt\relax}
\providecommand{\BIBentryALTinterwordstretchfactor}{4}
\providecommand{\BIBentryALTinterwordspacing}{\spaceskip=\fontdimen2\font plus
\BIBentryALTinterwordstretchfactor\fontdimen3\font minus \fontdimen4\font\relax}
\providecommand{\BIBforeignlanguage}[2]{{%
\expandafter\ifx\csname l@#1\endcsname\relax
\typeout{** WARNING: IEEEtran.bst: No hyphenation pattern has been}%
\typeout{** loaded for the language `#1'. Using the pattern for}%
\typeout{** the default language instead.}%
\else
\language=\csname l@#1\endcsname
\fi
#2}}
\providecommand{\BIBdecl}{\relax}
\BIBdecl

\bibitem{1-games-1}
V.~Mnih, K.~Kavukcuoglu, D.~Silver, A.~Graves, I.~Antonoglou, D.~Wierstra, and M.~Riedmiller, ``Playing atari with deep reinforcement learning,'' \emph{arXiv preprint arXiv:1312.5602}, 2013.

\bibitem{1-games-2}
D.~Silver, A.~Huang, C.~J. Maddison, A.~Guez, L.~Sifre, G.~Van Den~Driessche, J.~Schrittwieser, I.~Antonoglou, V.~Panneershelvam, M.~Lanctot \emph{et~al.}, ``Mastering the game of go with deep neural networks and tree search,'' \emph{nature}, vol. 529, no. 7587, pp. 484--489, 2016.

\bibitem{1-LLM-1}
L.~Ouyang, J.~Wu, X.~Jiang, D.~Almeida, C.~Wainwright, P.~Mishkin, C.~Zhang, S.~Agarwal, K.~Slama, A.~Ray \emph{et~al.}, ``Training language models to follow instructions with human feedback,'' \emph{Advances in Neural Information Processing Systems}, vol.~35, pp. 27\,730--27\,744, 2022.

\bibitem{1-LLM-2}
T.~Carta, C.~Romac, T.~Wolf, S.~Lamprier, O.~Sigaud, and P.-Y. Oudeyer, ``Grounding large language models in interactive environments with online reinforcement learning,'' \emph{arXiv preprint arXiv:2302.02662}, 2023.

\bibitem{1-robot-1}
Y.~Duan, X.~Chen, R.~Houthooft, J.~Schulman, and P.~Abbeel, ``Benchmarking deep reinforcement learning for continuous control,'' in \emph{International conference on machine learning}.\hskip 1em plus 0.5em minus 0.4em\relax PMLR, 2016, pp. 1329--1338.

\bibitem{1-robot-2}
J.~Hwangbo, I.~Sa, R.~Siegwart, and M.~Hutter, ``Control of a quadrotor with reinforcement learning,'' \emph{IEEE Robotics and Automation Letters}, vol.~2, no.~4, pp. 2096--2103, 2017.

\bibitem{1-remote-1}
T.~Lampe, L.~D. Fiederer, M.~Voelker, A.~Knorr, M.~Riedmiller, and T.~Ball, ``A brain-computer interface for high-level remote control of an autonomous, reinforcement-learning-based robotic system for reaching and grasping,'' in \emph{Proceedings of the 19th international conference on Intelligent User Interfaces}, 2014, pp. 83--88.

\bibitem{1-commu-1}
S.~B. Moon, P.~Skelly, and D.~Towsley, ``Estimation and removal of clock skew from network delay measurements,'' in \emph{IEEE INFOCOM'99. Conference on Computer Communications. Proceedings. Eighteenth Annual Joint Conference of the IEEE Computer and Communications Societies. The Future is Now (Cat. No. 99CH36320)}, vol.~1.\hskip 1em plus 0.5em minus 0.4em\relax IEEE, 1999, pp. 227--234.

\bibitem{1-previous-1}
K.~Gu and S.-I. Niculescu, ``Survey on recent results in the stability and control of time-delay systems,'' \emph{J. Dyn. Sys., Meas., Control}, vol. 125, no.~2, pp. 158--165, 2003.

\bibitem{1-previous-2}
L.~Dugard and E.~I. Verriest, \emph{Stability and control of time-delay systems}.\hskip 1em plus 0.5em minus 0.4em\relax Springer, 1998, vol. 228.

\bibitem{2-modelfree-1}
K.~V. Katsikopoulos and S.~E. Engelbrecht, ``Markov decision processes with delays and asynchronous cost collection,'' \emph{IEEE transactions on automatic control}, vol.~48, no.~4, pp. 568--574, 2003.

\bibitem{2-modelfree-2}
S.~Nath, M.~Baranwal, and H.~Khadilkar, ``Revisiting state augmentation methods for reinforcement learning with stochastic delays,'' in \emph{Proceedings of the 30th ACM International Conference on Information \& Knowledge Management}, 2021, pp. 1346--1355.

\bibitem{2-modelfree-3}
S.~Ramstedt and C.~Pal, ``Real-time reinforcement learning,'' \emph{Advances in neural information processing systems}, vol.~32, 2019.

\bibitem{2-modelfree-4}
T.~Xiao, E.~Jang, D.~Kalashnikov, S.~Levine, J.~Ibarz, K.~Hausman, and A.~Herzog, ``Thinking while moving: Deep reinforcement learning with concurrent control,'' \emph{arXiv preprint arXiv:2004.06089}, 2020.

\bibitem{2-modelfree-5}
E.~Schuitema, L.~Bu{\c{s}}oniu, R.~Babu{\v{s}}ka, and P.~Jonker, ``Control delay in reinforcement learning for real-time dynamic systems: A memoryless approach,'' in \emph{2010 IEEE/RSJ International Conference on Intelligent Robots and Systems}.\hskip 1em plus 0.5em minus 0.4em\relax IEEE, 2010, pp. 3226--3231.

\bibitem{2-modelfree-6}
M.~Agarwal and V.~Aggarwal, ``Blind decision making: Reinforcement learning with delayed observations,'' \emph{Pattern Recognition Letters}, vol. 150, pp. 176--182, 2021.

\bibitem{2-modelfree-7}
Y.~Bouteiller, S.~Ramstedt, G.~Beltrame, C.~Pal, and J.~Binas, ``Reinforcement learning with random delays,'' in \emph{International conference on learning representations}, 2021.

\bibitem{2-modelbased-1}
T.~J. Walsh, A.~Nouri, L.~Li, and M.~L. Littman, ``Planning and learning in environments with delayed feedback,'' in \emph{Machine Learning: ECML 2007: 18th European Conference on Machine Learning, Warsaw, Poland, September 17-21, 2007. Proceedings 18}.\hskip 1em plus 0.5em minus 0.4em\relax Springer, 2007, pp. 442--453.

\bibitem{2-modelbased-2}
T.~Hester and P.~Stone, ``Texplore: real-time sample-efficient reinforcement learning for robots,'' \emph{Machine learning}, vol.~90, pp. 385--429, 2013.

\bibitem{2-modelbased-3}
B.~Chen, M.~Xu, L.~Li, and D.~Zhao, ``Delay-aware model-based reinforcement learning for continuous control,'' \emph{Neurocomputing}, vol. 450, pp. 119--128, 2021.

\bibitem{2-modelbased-4}
V.~Firoiu, T.~Ju, and J.~Tenenbaum, ``At human speed: Deep reinforcement learning with action delay,'' \emph{arXiv preprint arXiv:1810.07286}, 2018.

\bibitem{2-modelbased-5}
E.~Derman, G.~Dalal, and S.~Mannor, ``Acting in delayed environments with non-stationary markov policies,'' \emph{arXiv preprint arXiv:2101.11992}, 2021.

\bibitem{GRU}
J.~Chung, C.~Gulcehre, K.~Cho, and Y.~Bengio, ``Empirical evaluation of gated recurrent neural networks on sequence modeling,'' \emph{arXiv preprint arXiv:1412.3555}, 2014.

\bibitem{haarnoja2018soft}
T.~Haarnoja, A.~Zhou, P.~Abbeel, and S.~Levine, ``Soft actor-critic: Off-policy maximum entropy deep reinforcement learning with a stochastic actor,'' in \emph{International conference on machine learning}.\hskip 1em plus 0.5em minus 0.4em\relax PMLR, 2018, pp. 1861--1870.

\bibitem{1}
M.~Vecerik, T.~Hester, J.~Scholz, F.~Wang, O.~Pietquin, B.~Piot, N.~Heess, T.~Roth{\"o}rl, T.~Lampe, and M.~Riedmiller, ``Leveraging demonstrations for deep reinforcement learning on robotics problems with sparse rewards,'' \emph{arXiv preprint arXiv:1707.08817}, 2017.

\bibitem{2}
T.~Hester, M.~Vecerik, O.~Pietquin, M.~Lanctot, T.~Schaul, B.~Piot, D.~Horgan, J.~Quan, A.~Sendonaris, I.~Osband \emph{et~al.}, ``Deep q-learning from demonstrations,'' in \emph{Proceedings of the AAAI Conference on Artificial Intelligence}, vol.~32, no.~1, 2018.

\bibitem{3}
S.~Lee, Y.~Seo, K.~Lee, P.~Abbeel, and J.~Shin, ``Offline-to-online reinforcement learning via balanced replay and pessimistic q-ensemble,'' in \emph{Conference on Robot Learning}.\hskip 1em plus 0.5em minus 0.4em\relax PMLR, 2022, pp. 1702--1712.

\bibitem{4}
Y.~Mao, C.~Wang, B.~Wang, and C.~Zhang, ``Moore: Model-based offline-to-online reinforcement learning,'' \emph{arXiv preprint arXiv:2201.10070}, 2022.

\bibitem{5}
P.~J. Ball, L.~Smith, I.~Kostrikov, and S.~Levine, ``Efficient online reinforcement learning with offline data,'' \emph{arXiv preprint arXiv:2302.02948}, 2023.

\bibitem{6}
A.~Nair, B.~McGrew, M.~Andrychowicz, W.~Zaremba, and P.~Abbeel, ``Overcoming exploration in reinforcement learning with demonstrations,'' in \emph{2018 IEEE international conference on robotics and automation (ICRA)}.\hskip 1em plus 0.5em minus 0.4em\relax IEEE, 2018, pp. 6292--6299.

\bibitem{7}
N.~Hansen, Y.~Lin, H.~Su, X.~Wang, V.~Kumar, and A.~Rajeswaran, ``Modem: Accelerating visual model-based reinforcement learning with demonstrations,'' \emph{arXiv preprint arXiv:2212.05698}, 2022.

\bibitem{8}
M.~Yang and O.~Nachum, ``Representation matters: offline pretraining for sequential decision making,'' in \emph{International Conference on Machine Learning}.\hskip 1em plus 0.5em minus 0.4em\relax PMLR, 2021, pp. 11\,784--11\,794.

\bibitem{9}
A.~Rajeswaran, V.~Kumar, A.~Gupta, G.~Vezzani, J.~Schulman, E.~Todorov, and S.~Levine, ``Learning complex dexterous manipulation with deep reinforcement learning and demonstrations,'' \emph{arXiv preprint arXiv:1709.10087}, 2017.

\bibitem{10}
A.~Nair, A.~Gupta, M.~Dalal, and S.~Levine, ``Awac: Accelerating online reinforcement learning with offline datasets,'' \emph{arXiv preprint arXiv:2006.09359}, 2020.

\bibitem{11}
Y.~Zhao, R.~Boney, A.~Ilin, J.~Kannala, and J.~Pajarinen, ``Adaptive behavior cloning regularization for stable offline-to-online reinforcement learning,'' \emph{arXiv preprint arXiv:2210.13846}, 2022.

\bibitem{12}
T.~G. Rudner, C.~Lu, M.~A. Osborne, Y.~Gal, and Y.~Teh, ``On pathologies in kl-regularized reinforcement learning from expert demonstrations,'' \emph{Advances in Neural Information Processing Systems}, vol.~34, pp. 28\,376--28\,389, 2021.

\bibitem{13}
I.~Uchendu, T.~Xiao, Y.~Lu, B.~Zhu, M.~Yan, J.~Simon, M.~Bennice, C.~Fu, C.~Ma, J.~Jiao \emph{et~al.}, ``Jump-start reinforcement learning,'' \emph{arXiv preprint arXiv:2204.02372}, 2022.

\bibitem{2-rl4rec-1}
M.~Chen, A.~Beutel, P.~Covington, S.~Jain, F.~Belletti, and E.~H. Chi, ``Top-k off-policy correction for a reinforce recommender system,'' in \emph{Proceedings of the Twelfth ACM International Conference on Web Search and Data Mining}, 2019, pp. 456--464.

\bibitem{2-rl4rec-2}
X.~Zhao, L.~Zhang, Z.~Ding, L.~Xia, J.~Tang, and D.~Yin, ``Recommendations with negative feedback via pairwise deep reinforcement learning,'' in \emph{Proceedings of the 24th ACM SIGKDD International Conference on Knowledge Discovery \& Data Mining}, 2018, pp. 1040--1048.

\bibitem{2-rl4rec-3}
F.~Liu, R.~Tang, X.~Li, W.~Zhang, Y.~Ye, H.~Chen, H.~Guo, Y.~Zhang, and X.~He, ``State representation modeling for deep reinforcement learning based recommendation,'' \emph{Knowledge-Based Systems}, vol. 205, p. 106170, 2020.

\bibitem{2-visual-1}
R.~Shah and V.~Kumar, ``Rrl: Resnet as representation for reinforcement learning,'' \emph{arXiv preprint arXiv:2107.03380}, 2021.

\bibitem{2-visual-2}
S.~Parisi, A.~Rajeswaran, S.~Purushwalkam, and A.~Gupta, ``The unsurprising effectiveness of pre-trained vision models for control,'' in \emph{International Conference on Machine Learning}.\hskip 1em plus 0.5em minus 0.4em\relax PMLR, 2022, pp. 17\,359--17\,371.

\bibitem{2-visual-3}
Z.~Yuan, Z.~Xue, B.~Yuan, X.~Wang, Y.~Wu, Y.~Gao, and H.~Xu, ``Pre-trained image encoder for generalizable visual reinforcement learning,'' \emph{arXiv preprint arXiv:2212.08860}, 2022.

\bibitem{2-marl-1}
H.~Ge, D.~Gao, L.~Sun, Y.~Hou, C.~Yu, Y.~Wang, and G.~Tan, ``Multi-agent transfer reinforcement learning with multi-view encoder for adaptive traffic signal control,'' \emph{IEEE Transactions on Intelligent Transportation Systems}, vol.~23, no.~8, pp. 12\,572--12\,587, 2021.

\bibitem{RDDP-1}
K.~V. Katsikopoulos and S.~E. Engelbrecht, ``Markov decision processes with delays and asynchronous cost collection,'' \emph{IEEE transactions on automatic control}, vol.~48, no.~4, pp. 568--574, 2003.

\bibitem{RDDP-2}
S.~Nath, M.~Baranwal, and H.~Khadilkar, ``Revisiting state augmentation methods for reinforcement learning with stochastic delays,'' in \emph{Proceedings of the 30th ACM International Conference on Information \& Knowledge Management}, 2021, pp. 1346--1355.

\bibitem{seq2seq}
I.~Sutskever, O.~Vinyals, and Q.~V. Le, ``Sequence to sequence learning with neural networks,'' \emph{Advances in neural information processing systems}, vol.~27, 2014.

\bibitem{zhang2023policy}
H.~Zhang, W.~Xu, and H.~Yu, ``Policy expansion for bridging offline-to-online reinforcement learning,'' \emph{arXiv preprint arXiv:2302.00935}, 2023.

\bibitem{CDMDP}
T.~J. Walsh, A.~Nouri, L.~Li, and M.~L. Littman, ``Learning and planning in environments with delayed feedback,'' \emph{Autonomous Agents and Multi-Agent Systems}, vol.~18, pp. 83--105, 2009.

\bibitem{franceschetti2022making}
M.~Franceschetti, C.~Lacoux, R.~Ohouens, and O.~Sigaud, ``Making reinforcement learning work on swimmer,'' \emph{arXiv preprint arXiv:2208.07587}, 2022.

\end{thebibliography}


\appendices

\section{MORE DISCUSSION}

\subsection{Explanation on interpretability}
\label{interpretability}

Information state is employed to address delayed RL problems, comprising the latest observations of the agent and all action sequences from the moment of self-observation to the current time. 
When using the Information State directly in RL algorithms, two issues arise: 1) As the delay increases, the proportion of state in the Information State decreases.
For example, in the HalfCheetah task, with a state space dimension of 17 and action space dimension of 6, the proportion of the state part and action part is only equal when the delay is 3. 
As the delay increases, the proportion of the action part becomes larger. Consequently, when making decisions directly using RL algorithms, neural networks are more likely to be influenced by the part with a higher proportion, affecting decision-making effectiveness. 
2) Directly training end-to-end using the information state is a black-box process, where the final policy performance depends on the ability of the neural network used for training, particularly its network structure, capacity, and parameter settings.

DEER maps the information state to a latent space and makes decisions based on the representation vectors, regardless of whether the current task is fixed or random. 
Compared to end-to-end training, DEER is more directional because the training process of its encoder enables it to better map the Information State to a representation closer to the delayed process state. 
Consequently, the encoded state is more conducive to decision-making.
The entire process can be likened to a student choosing the shortest route from home to school. 
End-to-end learning is akin to starting without a purpose, recording the time taken each time, and then selecting points of travel with less time consumption through comparison, gradually adjusting to the most suitable route. 
DEER, on the other hand, is similar to knowing from the start that there is a place closer to school, and we know the route to this place. 
We first reach this place directly and then use an end-to-end approach to find the shorter routes thereafter.
The improved interpretability of DEER is reflected in this aspect.

\subsection{Difference between DEER and model-based methods}
\label{robustness}
Model-based algorithms typically entail constructing a dynamics model through real-time interactions with the environment, followed by planning using this model. 
Consequently, these algorithms are often susceptible to model and iteration errors. 
In contrast, DEER utilizes pre-existing data from delay-free environments to train an encoder-decoder for multi-step prediction.
During the decision-making phase, DEER utilizes the encoder's representation of delayed states and their corresponding action sequences, referred to as context representations, instead of the final states from multi-step predictions. 
This representation extracts decision-relevant information from the most recently observed states and action sequences, thereby mitigating sensitivity to deviations from the true state caused by model and iteration errors.

\subsection{Difference between RDMDP and RDDMDP}
\label{appendix_discussion_1}
The RDMDP proposed in \cite{RDDP-1} and \cite{RDDP-2} introduces stochasticity by considering variable time steps between two successive variables, including observations, actions and rewards. The value of these time steps is not fixed and can differ from unity. Additionally, the observation of a later state $s_{t+1}$ can only occur after the observation of the current state $s_t$. This implies that the state assumed to be observed at time $t$ is actually observed at a later time $t+d$, where $d$ represents the accumulated delay values before time $t$ and it gradually increases over time. However, the RDDMDP we study extends the constant delay scenario by introducing a bounded number of steps with missing states. In our settings, if the state $s_{t-d_I}$ was observed at time $t$, the next state and reward would be $s_{t-d_I+1}$ and $r_{t-d_I+1}$ with probability $1 - \mu$ while such information could not be acquired with probability $ \mu$ and we would replace them with $s_{t-d_I}$ and $r_{t-d_I}$. Therefore, in this paper, the random delay value is defined as the number of time steps between two observed  states. If the next state received in RDDMDP precisely corresponded to the state at the subsequent moment in time, RDDMDP would be equivalent to RDMDP.

\subsection{Constant Delayed Markov Decision Process (CDMDP)}
\label{CDMDP}

The dropping probabilty $\mu = 0$ means that there is no information dropout during the interaction between the environment and the agent. Consequently, the RDDMDP can simplified as the Constant Delayed MDP (CDMDP) \cite{CDMDP}.
\begin{definition}
     Similar to RDDMDP $(d_I, d_M, \bm{\mathcal{I}_z},\bm{\mathcal{A}},\bm{\rho},\bm{p},\bm{r},\gamma,\mu)$, CDMDP can be defined as a 6-tuple $(\bm{\mathcal{I}_d},\bm{\mathcal{A}},\bm{\rho},\bm{p},\bm{r},\gamma)$:

(1) Information state space:  $\bm{\mathcal{I}_d}=\mathcal{S}\times\mathcal{A}^d$, 
where $d$ denotes the delay step and $\bm{i}_t=(s_{t-d},(a_{t-n}^{(t)})_{n=d:1}) \in \bm{\mathcal{I}_d}$;

(2) Action space:  $\bm{\mathcal{A}}=\mathcal{A}$;

(3) Initial information state distribution: 

$\bm{\rho}(\bm{i}_0)=\bm{\rho}(s_0,a_0,...,a_{d-1})=\rho(s_0)\prod_{i=0}^{d-1}\delta(a_i-c_i)$, 

\noindent where ${(c_i)}_{i=0:d-1}$ denotes the initial action sequence and $\delta$ is the Dirac delta function;

(4) Transition distribution: \\ 
$\bm{p}(\bm{i}_{t+1}|\bm{i}_{t},\bm{a}_t)\\
=\bm{p}(s_{t-d+1},a_{t-d+1}^{(t+1)},...,a_{t}^{(t+1)}|s_{t-d},a_{t-d}^{(t)},...,a_{t-1}^{(t)},\bm{a}_t)\\
=p(s_{t-d+1}|s_{t-d},a_{t-d})\prod_{i=1}^{d-1}\delta(a_{t-d+i}^{(t+1)}-a_{t-d+i}^{(t)})\delta(a_{t}^{(t+1)}-\bm{a}_t)$;

(5) Reward function: $\bm{r}_t=r_{t-d}$;

(6) Discount factor: $\gamma \in [0,1)$.

\end{definition}

\subsection{A concise example}
\label{demonstration}
\begin{figure}[htbp]
\includegraphics[width=0.49\textwidth]{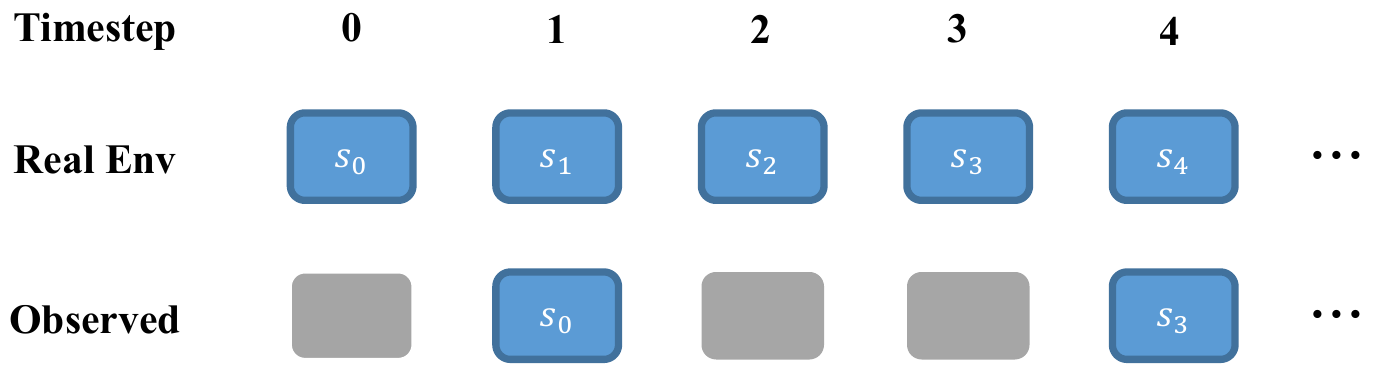}
\centering
\caption{A simple demonstration of the RDDMDP. Gray squares represent information dropout, while blue squares indicate received information.}
 \label{pic:demonstration}
\end{figure}

A concise example of the RDDMDP is provided in Figure \ref{pic:demonstration}. 
Assuming an intrinsic delay $d_I=1$ and a maximum number of extra dropping steps $d_M=1$, this leads to a random delay value $1 \le z \le 2$.
The agent makes decisions based on the feedback from the delayed environment, as outlined in Table \ref{tab:process}.

\begin{table}[htbp]
  \centering
  \caption{Process of decision-making in delayed environment}
  \renewcommand\arraystretch{1.2}
  \scalebox{0.9}{
    \begin{tabular}{ccccc}
    \toprule
    \bf Timestep & \bf Observation &  \bf Random Value ($z$) &  \bf Information State & \bf Action \\
    \midrule
     \bf 0  & None & 1 & None & $a_0$ \\
     \bf 1  & $s_0$ & 1 & ($s_0$, $a_0$) & $a_1$ \\
     \bf 2  & None & 2 & ($s_0$, $a_0$, $a_1$) & $a_2$ \\
     \bf 3  & None & 2 & ($s_0$, $a_1$, $a_2$) & $a_3$ \\
     \bf 4  & $s_3$ & 1 & ($s_3$, $a_3$) & $a_4$ \\
    \bottomrule
    \end{tabular}%
  }
  \label{tab:process}%
\end{table}%

\subsection{Analysis of results in Swimmer using DEER}
\label{Swimmer Analysis}
As depicted in Figure \ref{comparision with constant delay}, DEER (delay=0) consistently outperforms expert performance across all environments, with a notable performance boost in the Swimmer environment, achieving nearly 2.5 times higher returns. This implies that DEER exhibits an elevated upper limit for task completion in environments with diverse delays.
Additionally, as discussed in the article \cite{franceschetti2022making}, the Swimmer environment presents a unique challenge, where traditional RL methods often converge to a local optimum around a score of 40, whereas direct policy search algorithms can achieve scores close to 300. 
This discrepancy doesn't arise from the incapacity of RL methods to discover high-scoring Snake-like behaviors but rather results from the presence of the discount factor, diminishing the impact of later steps in cumulative reward.

In our experiments, we set the discount factor $\gamma$ to 0.99 for all tasks. It happens that $0.99^{1000}=0.00004317124$, thus the contribution of the last step compared to the first step in cumulative reward is only $4.32 \times 10^{-5}$. 
To address this issue, the author proposes setting the discount factor to 1, illustrating that in the Swimmer task, each state holds an equal impact on the final task performance.
While DEER maintains the setting of $\gamma=0.99$, it achieves accumulated returns that are 2.5 times higher than those of the RL expert. This is attributed to the fact that the context representation in DEER encompasses all delayed state information, as opposed to a single-state representation, leading to more effective decision-making.

\section{IMPLEMENTATION DETAILS}
\subsection{Details of the pretrained model}
\label{appendix_implementaion_encoder}

\begin{figure*}[t]
    \centering
    \includegraphics[width=0.85\textwidth]{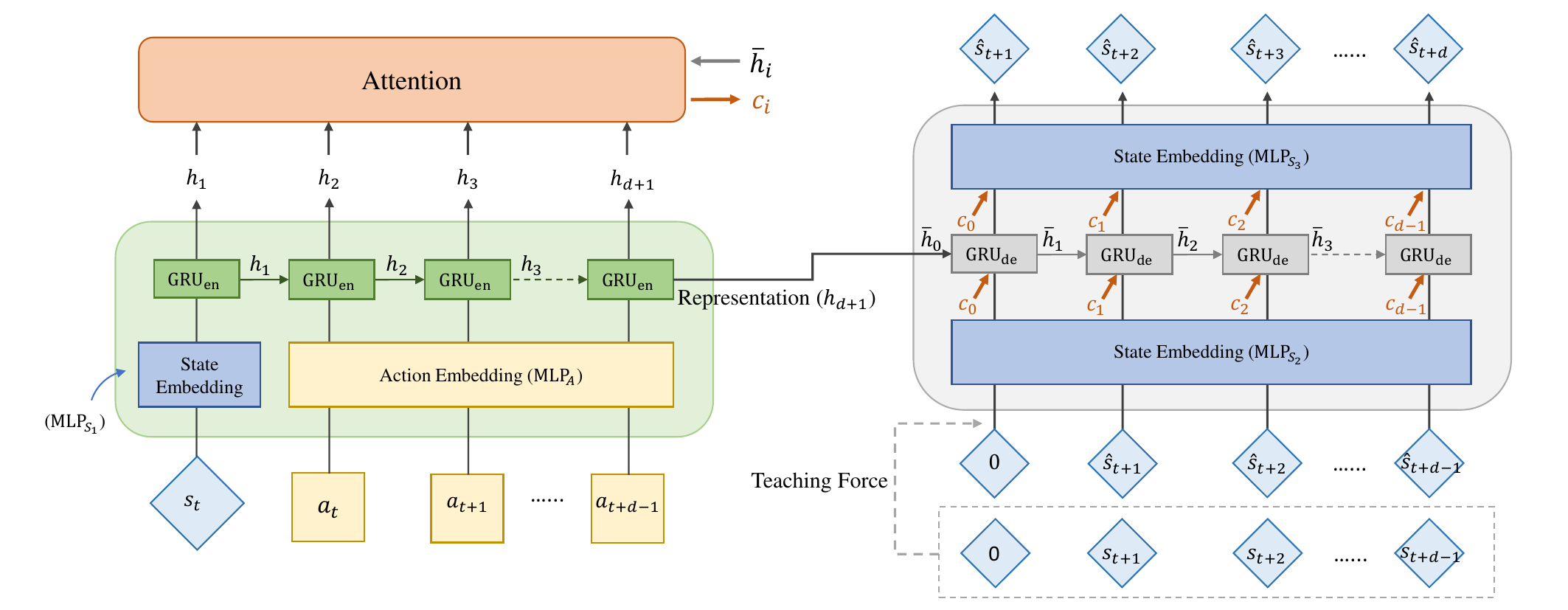}
    \caption{Network structure of pretrained model. It comprises two key modules: the encoder module and the decoder module. In the encoder module, the context representation is generated by first encoding the state and the action sequence within the information state using $\text{MLP}_{S_1}$ and $\text{MLP}_{A}$, respectively. The resulting encodings are then fed into the $\text{GRU}_\text{en}$ to obtain hidden states. The decoder module, on the other hand, is responsible for restoring the encoded information. The state sequence is processed by $\text{MLP}_{S_2}$, followed by inputting it into $\text{GRU}_\text{de}$ along with the attention $c_i$ and hidden state $\bar{h}_i$. The final states are achieved by applying $\text{MLP}_{S_3}$. Teaching Force is employed within the decoder to enhance training efficiency.
    }
    \label{encoder_structure}
\end{figure*}

In this section, we present the detailed settings for the Encoder. The complete structure of the pretrained model is shown in Fig.\ref{encoder_structure}. Particularly, in the encoder module, the relevant state representation is computed as follows:
\begin{equation*}
    \begin{aligned}
        h_1&=\text{GRU}_\text{en}(\text{MLP}_{S_1}(s_t)) ,\\
   h_i&=\text{GRU}_\text{en}(\text{MLP}_A(a_{t+i-2}), h_{i-1}),
    \end{aligned}
\end{equation*}
where $i=2,3,...,d+1$, $h_{d+1}$ is the representation utilized in the policy training phase and the subscript $d$ denotes the delay value. In the decoder, we use teacher forcing and attention to strengthen the performance,
\begin{equation*}
    \begin{aligned}
        \bar{h}_0&=h_{d+1}, \\
      c_i&=\text{Attention}((h_1,...,h_{d+1}),\bar{h}_i), \\
      \bar{h}_1&=\text{GRU}_\text{de}(\text{MLP}_{S_2}(0)\oplus c_0,\bar{h}_{0}),
    \end{aligned}
\end{equation*}
\begin{equation*}
    \begin{aligned}
      \bar{h}_i=
      \begin{cases} \text{GRU}_\text{de}(\text{MLP}_{S_2}(\hat{s}_{t+i-1})\oplus c_{i-1},\bar{h}_{i-1}), \\ \quad\quad\quad\quad \text{with probability } p \ , \\ 
    \text{GRU}_\text{de}(\text{MLP}_{S_2}(s_{t+i-1})\oplus c_{i-1},\bar{h}_{i-1}),  \\ \quad\quad\quad\quad \text{with probability } 1-p   ,
    \end{cases} i=2,3,...,d, 
    \end{aligned}
\end{equation*}
\begin{equation*}
    \begin{aligned}
    \hat{s}_{t+i}=\text{MLP}_{S_3}(\bar{h}_i\oplus c_{i-1}),\ i=1,2,...,d,
    \end{aligned}
\end{equation*}
where $p$ denotes the teacher forcing ratio and $\oplus$ denotes the concatenation of tensor along the last dimension.

The parameters of the pretrained model are shown in Table \ref{parameters}. $\text{K}_1$ and $\text{K}_2$ are both hyperparameters, which denote the dimension of hidden states in GRU and the dimension of embeddings in MLP separately. In this paper, we conduct experiments with $\text{K}_1$=128, 256, 512, respectively with $\text{K}_2$ = 64.

\begin{table}[htbp]
  \centering
  \caption{Parameters of the pretrained model}
  \renewcommand\arraystretch{1.2}
    \begin{tabular}{ccc}
    \toprule
    {\bf Name}  & {\bf Number of Layer}  & {\bf Parameters} \\
    \midrule
    $\text{GRU}_\text{en}$         & 1 Layer & ($\text{K}_2$, $\text{K}_1$) \\
$\text{GRU}_\text{de}$         & 1 Layer & ($\text{K}_1$ + $\text{K}_2$, $\text{K}_1$) \\
$\text{MLP}_{S_1}$         & 1 Layer & (state dimension, $\text{K}_2$) \\
$\text{MLP}_{A}$         & 1 Layer & (action dimension, $\text{K}_2$) \\
$\text{MLP}_{S_2}$         & 1 Layer & (state dimension, $\text{K}_2$) \\
$\text{MLP}_{S_3}$         & 1 Layer & ($2$$\text{K}_1$, state dimension) \\
    \bottomrule
    \end{tabular}%
  \label{parameters}%
\end{table}%

\subsection{Pseudocode}
The pseudocode for DEER is shown in Algorithm \ref{algorithm}.

\begin{algorithm*}[!h]
\caption{Delay-resilient Encoder-Enhanced Reinforcement Learning(DEER)}
\label{algorithm}
\renewcommand\arraystretch{1.2}
\begin{algorithmic}[1]
\Require $M$ random trajectories, $N$ expert trajectories$ (M >> N)$, maximum number of delayed steps $D$.
\Ensure Policy $\pi_{\theta}$ in delayed environment.

\State \textbf{Stage 1(Pretraining Model):} 
\State Convert $M+N$ trajectories into training and test datasets, following the data format: $(s_{t-D}, a_{t-D}, \cdots, a_{t-1})$ and corresponding labels: $(s_{t-D+1}, ..., s_{t})$, where datasets contain various delays from 1 to $D$ and action sequence in the information state is padded with zeros until its length matches $D$.
\State Initialize a Seq2Seq Model comprising $Encoder(\cdot)$ and $Decoder(\cdot)$.
\State Train the Seq2Seq Model using supervised learning on the constructed dataset.
\State Output the $Encoder(\cdot)$ from the trained Seq2Seq Model.
\State
\State \textbf{Stage 2 (Decision Model):} 
\For{each iteration}
 \For{each environment step}
   \State Identify current delayed step $d$.
   \State Obtain the information state $I_t = (s_{t-d}, a_{t-d}, ..., a_{t-1})$.
   \State Compute context\_representation $h_t = Encoder(I_t)$.
   \State Take action $a_t = \pi_{\theta}(h_t)$ in the environment, obtain next state $s_{t-d+1}$, and reward $\bm{r}_t$.
   \State Calculate context\_representation $h_{t+1} = Encoder(I_{t+1})$.
   \State Store the transition $(h_t, a_t, \bm{r}_t, h_{t+1})$ in the replay buffer $\bm{R}$.
   \If{$len(\bm{R}) \geq training\_threshold$}
     \State Update policy $\pi_{\theta}$ using the Soft Actor-Critic (SAC) method.
   \EndIf
 \EndFor
\EndFor
\State Output policy $\pi_{\theta}$ in delayed environments.
\end{algorithmic}
\end{algorithm*}

\newpage

\subsection{Datasets for the pretraining process}
\label{offline datasets}

The dataset used for pretraining consists of two distinct components: random trajectories and expert trajectories. 
Random trajectories are generated by allowing the agent to take random actions. 
Expert trajectories are produced through a two-step process: 
first, an agent is trained using the Soft Actor-Critic (SAC) algorithm in a delay-free environment to achieve proficient task performance; 
second, this trained agent is utilized to generate the required number of expert trajectories that meet the specified conditions.

To better simulate real-world scenarios, DEER pre-trains the encoder using a majority of random trajectories combined with a minority of expert trajectories, as shown in the "Mimicking Reality" section of Table \ref{tab: offline dataset}. 
Additionally, to assess the impact of offline datasets on the model, two more datasets are introduced for comparison: one consisting entirely of expert trajectories and the other entirely of random trajectories. 
To ensure an equal data volume across all three datasets, the number of trajectories for the newly introduced datasets is provided in other sections in Table \ref{tab: offline dataset}.

\begin{table*}[t]
  \centering
  \caption{Datasets for pretraining the Seq2Seq model }
  \renewcommand\arraystretch{1.2}
    \begin{tabular}{c|cccccc}
    \toprule
    \bf Case  & \multicolumn{6}{c}{\bf Mimicing Reality} \\
    \midrule
   \bf Env & \bf Ant & \bf HalfCheetah & \bf Hopper & \bf Swimmer & \bf Walker2d & \bf Reacher
    \\
    \midrule
    \bf Random & 100 & 500 & 800 & 500 & 8000 & 2000 \\
    \bf Expert & 10 & 10 & 10 & 10 & 60 & 10 \\
    \midrule
    \midrule
    \bf Case  & \multicolumn{6}{c}{\bf Entirely Random} \\
    \midrule
   \bf Env & \bf Ant & \bf HalfCheetah & \bf Hopper & \bf Swimmer & \bf Walker2d & \bf Reacher
    \\
    \midrule
    \bf Random & 200 & 600 & 1000 & 600 & 9000 & 2100 \\
    \bf Expert & 0 & 0 & 0 & 0 & 0 & 0 \\
    \midrule
    \midrule
    \bf Case  & \multicolumn{6}{c}{\bf Entirely Expert} \\
    \midrule
   \bf Env & \bf Ant & \bf HalfCheetah & \bf Hopper & \bf Swimmer & \bf Walker2d & \bf Reacher
    \\
    \midrule
    \bf Random & 0 & 0 & 0 & 0 & 0 & 0 \\
    \bf Expert & 100 & 300 & 400 & 300 & 800 & 500 \\
    \bottomrule
    \end{tabular}%
  \label{tab: offline dataset}%
\end{table*}%

\newpage

\end{document}